%% file: m7106.tex
\documentclass{ecai} 
\usepackage{latexsym}
\usepackage{amssymb}
\usepackage{amsmath}
\usepackage{amsthm}
\usepackage{booktabs}
\usepackage{enumitem}
\usepackage{graphicx}
\usepackage{color}
\usepackage{multibib}
\usepackage{multirow}
\usepackage{booktabs}
\usepackage{caption}
\usepackage{enumitem}
\usepackage{stfloats}
\usepackage[noend]{algpseudocode}
\usepackage[linesnumbered,vlined,ruled,commentsnumbered]{algorithm2e}
\newtheorem{theorem}{Theorem}

\newtheorem{proposition}[theorem]{Proposition}

\newcommand{\BibTeX}{B\kern-.05em{\sc i\kern-.025em b}\kern-.08em\TeX}

\begin{document}

%%%%%%%%%%%%%%%%%%%%%%%%%%%%%%%%%%%%%%%%%%%%%%%%%%%%%%%%%%%%%%%%%%%%%%%%

\begin{frontmatter}

%%% Use this command to specify your submission number.
%%% In doubleblind mode, it will be printed on the first page.

\paperid{7106} 

%%% Use this command to specify the title of your paper.

\title{BadVim: Unveiling Backdoor Threats in \\ Visual State Space Model}

%%% Use this combinations of commands to specify all authors of your 
%%% paper. Use \fnms{} and \snm{} to indicate everyone's first names 
%%% and surname. This will help the publisher with indexing the 
%%% proceedings. Please use a reasonable approximation in case your 
%%% name does not neatly split into "first names" and "surname".
%%% Specifying your ORCID digital identifier is optional. 
%%% Use the \thanks{} command to indicate one or more corresponding 
%%% authors and their email address(es). If so desired, you can specify
%%% author contributions using the \footnote{} command.
% \footnote{Equal contribution.} \orcid{0009-0001-5790-2043}
\author[A]{\fnms{Cheng-Yi}~\snm{Lee}}
% \orcid{0009-0001-5790-2043}
\author[B]{\fnms{Yu-Hsuan}~\snm{Chiang}}
\author[C]{\fnms{Zhong-You}~\snm{Wu}} 
\author[C]{\fnms{Chia-Mu}~\snm{Yu}} 
% \orcid{0000-0002-1677-2131}
\author[A]{\fnms{Chun-Shien}~\snm{Lu}\thanks{Corresponding Author. Email: lcs@iis.sinica.edu.tw}}
% \orcid{0000-0002-5900-0019}
\address[A]{Academia Sinica}
\address[B]{National Central University}
\address[C]{National Yang Ming Chiao Tung University}

%%% Use this environment to include an abstract of your paper.

\begin{abstract}
Visual State Space Models (VSSM) have shown remarkable performance in various computer vision tasks. However, backdoor attacks pose significant security challenges, causing compromised models to predict target labels when specific triggers are present while maintaining normal behavior on benign samples. In this paper, we investigate the robustness of VSSMs against backdoor attacks. Specifically, we delicately design a novel framework for VSSMs, dubbed BadVim, which utilizes low-rank perturbations on state-wise to uncover their impact on state transitions during training. By poisoning only $0.3\%$ of the training data, our attacks cause any trigger-embedded input to be misclassified to the targeted class with a high attack success rate (over $97$\%) at inference time. Our findings suggest that the state-space representation property of VSSMs, which enhances model capability, may also contribute to its vulnerability to backdoor attacks. Our attack exhibits effectiveness across three datasets, even bypassing state-of-the-art defenses against such attacks. Extensive experiments show that the backdoor robustness of VSSMs is comparable to that of Transformers (ViTs) and superior to that of Convolutional Neural Networks (CNNs). We believe our findings will prompt the community to reconsider the trade-offs between performance and robustness in model design.
\end{abstract}

\end{frontmatter}

%%%%%%%%%%%%%%%%%%%%%%%%%%%%%%%%%%%%%%%%%%%%%%%%%%%%%%%%%%%%%%%%%%%%%%%%

\input{sec/1_intro}
\input{sec/2_related}

\input{sec/3_pre}
\input{sec/4_ours}

\input{sec/5_expt}
\input{sec/6_discuss}
\input{sec/7_conclusion}

\clearpage
%%% Use this environment to include acknowledgements (optional).
%%% This will be omitted in double-blind mode.
\begin{ack}
\vspace{-0.05in}
This work was supported by the National Science and Technology Council (NSTC) with Grant NSTC 112-2221-E-001-011-MY2 and Academia Sinica with Grant AS-IAIA-114-M08. 
\end{ack}

% We thank to National Center for High-performance Computing (NCHC) for providing computational and storage resources.
% By using the \texttt{ack} environment to insert your (optional) 
% acknowledgements, you can ensure that the text is suppressed whenever you use the \texttt{doubleblind} option. In the final version, acknowledgements may be included on the extra page intended for references.

% \vspace{-0.05in}
%%%%%%%%%%%%%%%%%%%%%%%%%%%%%%%%%%%%%%%%%%%%%%%%%%%%%%%%%%%%%%%%%%%%%%%%

%%% Use this command to include your bibliography file.

{\small
\bibliography{reference}
}

\clearpage
% \appendix

\input{sec/X_suppl}

\end{document}

%% file: sec/1_intro.tex
\section{Introduction}
\label{sec:intro}

% lenz2025jamba
State space models (SSMs)~\citep{gu2022efficiently, smithsimplified, fuhungry} have shown remarkable success in sequence modeling by capturing long-range dependencies and temporal dynamics. Unlike prevailing Transformer models with quadratic complexity, the state-space-based Mamba~\citep{gu2023mamba,dao2024transformers} achieves effective sequence modeling with linear complexity. This advantage has driven the widespread adoption of Mamba in large language models~\citep{wang2024the}, accelerating training and inference while significantly enhancing computational efficiency and scalability, as well as in visual recognition tasks. Recent studies show that visual state space models (VSSMs)~\citep{zhu2024vision,liu2024vmamba,hatamizadeh2024mambavision} deliver competitive performance while maintaining lower latency and memory overhead, highlighting their effectiveness in low-level vision tasks~\citep{wu2024rainmamba}, segmentation~\citep{xing2024segmamba}, and 3D object detection~\citep{zhang2024voxel}.

% zou2024freqmamba
% have gained prominence in vision tasks.
% Mamba~\citep{gu2023mamba,dao2024transformers}, an advanced SSM, offers improved SSMs efficiency and scalability, enhancing performance across various tasks. 

% Unlike vision transformers (ViTs) that rely on self-attention mechanisms or convolutional neural networks (CNNs) that prioritize local feature extraction, Vision Mamba exploits continuous-time dynamics to process visual data effectively.

% The VSS model has served as a generic and efficient backbone network designed to handle visual tasks, particularly through the advanced SSM mechanism, \textit{i.e.}, Mamba. Mamba addresses limitations in capturing sequence context and linear time complexity, enabling efficient interactions between sequential states. 

% Recent research: liang2024pointmamba, xiao2025spatialmamba

% Backdoor threat
% liu2018trojaning, vassilev2024adversarial
Unfortunately, deep neural networks are susceptible to various adversarial threats~\citep{papernot2016limitations}. One such threat, {\em i.e.}, backdoor attacks~\citep{gu2019badnets,li2022backdoor}, allows an adversary to manipulate a portion of the training data by injecting a trigger (\textit{i.e.}, a particular pattern), resulting in a backdoored model that entangles an incorrect correlation between the trigger patterns and target labels during the training process. In the inference stage, the backdoored model behaves normally with benign samples, but generates malicious and intended predictions when the trigger is activated. While prior work~\citep{du2024understanding,qi2024exploring} has examined the robustness of VSSMs against adversarial attacks, the vulnerability of VSSMs to backdoor attacks remains largely unexplored.

% While previous studies~\citep{du2024understanding} discuss the robustness of VSSMs to adversarial attacks, the issue of backdoor attacks remains largely unexplored in the VSSMs literature.

% We systematically compared with CNN to elucidate 

% backdoor/adversarial in mamba
% provide/contribute valuable insights/perspectives in future

% improves computational efficiency and enables effective long-range dependency modeling. low-rank properties while also maintaining stealthiness in VSS models?

\begin{figure}[!t]
    \centering
    \includegraphics[width=\columnwidth]{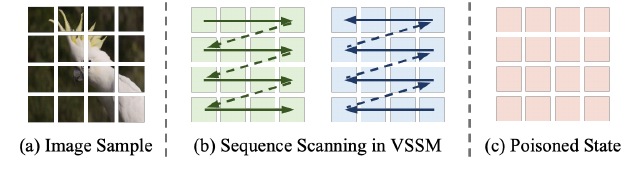}
    \vspace{0.1mm}
    \caption{Illustration of BadVim. (a) A clean input image is represented as a sequence of patches. (b) Sequence scanning in VSSMs (\textit{e.g.}, bidirectional scanning in Vim~\citep{zhu2024vision}). (c) BadVim injects triggers into each state before scanning, thereby ensuring the consistent implantation of the backdoor across the entire sequence.}
    \vspace{0.5cm}
    \label{fig:illust}
\end{figure}

% Our motivation/method introduction
To address this gap, we first revisit the distinct role of SSMs in the state transition matrix, \textit{i.e.}, the normal plus low-rank decomposition, which introduces local bias and positional information essential to the model. Building on this, we pose a question: ``\textit{Can we design an effective backdoor attack that disrupts the local bias components of this matrix while maintaining stealth in VSS models?}'' Accordingly, we introduce the BadVim framework, which reveals a new class of backdoor in VSSMs. At the core of BadVim is the Low-Rank Perturbation (LRP) trigger, which leverages the low-rank matrix structure to manipulate state transitions and reinforce the association between the trigger and target class. Given that SSMs update their internal state iteratively---each step influenced by the previous one---we argue that \textit{an effective trigger pattern should be embedded within each patch (state)} to ensure the consistent implantation of the backdoor, regardless of the model's state scanning trajectory. To better understand how perturbations influence SSMs, we present an analytical perspective on their effect on state transitions and model dynamics. 

We then empirically evaluate the robustness of VSSMs, vision transformers (ViTs), and convolutional neural networks (CNNs) under state-of-the-art (SOTA) backdoor attacks, revealing that VSSMs are generally more vulnerable to most triggers, similar to ViTs. In contrast, Gated CNNs~\citep{yu2025mambaout}, which lack the SSM module, demonstrate greater robustness against backdoor attacks. To further explore the vulnerability of VSSMs against our BadVim, we compare patch-based defenses (\textit{e.g.}, PatchDrop~\citep{naseer2021intriguing}, PatchShuffle~\citep{kang2017patchshuffle}) and two model-agnostic defenses~\citep{liu2018fine,zhu2023enhancing}. The results demonstrate that BadVim can successfully pass several backdoor defenses while maintaining superior accuracy on clean samples. We summarize our key contributions as follows:

% ($ii$) the lower bound on the sensitivity of SSMs to low-rank perturbations, which reflects the model's vulnerability, also sheds light on the minimum effective impact that low-rank triggers can have on the state transitions.

% Our study sheds light on critical aspects that are essential for further refinement.

% Currently, only  However, this study solely focuses on adversarial attacks (\textit{e.g.}, Projected Gradient Descent (PGD)~\citep{madry2018towards} and Patch-fool~\citep{fu2022patchfool}) rather than other security issues, \textit{e.g.}, backdoor attacks. It also compares the robustness of the VSS architecture with other architectures but does not specifically evaluate the robustness of the SSM module itself, \textit{i.e.}, its ability to resist attacks.  This approach limits the exploration of robustness in relation to the SSM module.

% The contributions of this paper are summarized below: 
% Our main contributions in this paper are three-fold.

% 主要論文貢獻在:
% 1. 探討SSM計算機制(Normal plus low-rank)，並提出後門攻擊
% 2. 使用定理解釋low-rank會對vision state space model為何有效
% 3. 經過實驗說明our trigger對vision state space model最有效，且比較不同模型架構(ResNet, Vision Transformer, GatedCNNs)說明vision state space model的robustness，提到GatedCNNs(無SSM機制)較為robust

\begin{itemize}
    \item We present a systematic investigation of VSSMs' vulnerability to backdoor attacks. By revisiting the unique role of the state transition matrix in SSMs, we identify a critical weakness in its low-rank components, unveiling an overlooked attack surface within the VSSM architecture.
 
    \item We propose BadVim, a novel backdoor attack framework tailored to VSSMs. BadVim introduces Low-Rank Perturbation (LRP) triggers that indirectly contaminate the state transition matrix, enhancing backdoor attack performance within the state-scanning mechanism of VSSMs.
    
    \item Experimental results demonstrate the effectiveness of BadVim across three datasets and evading existing defenses. Our findings show that the backdoor robustness of VSSMs is comparable to that of ViTs and stronger than that of CNNs.

    % \item We also evaluate patch-based and model-agnostic defenses, demonstrating the stealth and effectiveness of our BadViM attack. We show that our LRP successfully implants backdoors on VSS and circumvents state-of-the-art model-agnostic defenses.
\end{itemize}

% (1) We provide a comprehensive discussion about the success conditions of current main-stream backdoor defenses. We reveal that their success all relies on a prerequisite that backdoor triggers are sample-agnostic. 
% (2) We explore a novel invisible attack paradigm, where the backdoor trigger is sample-specific and invisible. It can bypass existing defenses for it breaks their fundamental assumption. 
% (3) Extensive experiments are conducted, which verify the effectiveness of the proposed method.

% \begin{itemize}
% \item We find that the SSM mechanism makes the VSS model more susceptible to backdoor attacks compared to the Gated CNN model, which does not incorporate SSM. We then examine the justification for this finding by analyzing the differences between these models.
% % backdoor robustness
% \item We observe that patch-processing defenses diminish the effectiveness of backdoor attacks in the VSS model. To counter this, we present a trigger pattern that recurs in each patch and exhibits a high success rate (ASR) even at a poisoning rate as low as $0.3$\%.
% \item We comprehensively evaluate the robustness of VSS model across three classical datasets and various backdoor attacks. Extensive experiments show that the backdoor robustness of VSS model is comparable to that of ViTs and superior to CNNs, except for Gated CNNs. % conduct experiments on three datasets and 
% \end{itemize}

%% file: sec/2_related.tex
\section{Related Works}
\subsection{Visual State Space Model}
% 簡述一下state space model
% 說明S4 model的缺點，以及S4 model related work 的介紹
Structured State-Space Sequence (S4)~\citep{gu2022efficiently}, a novel approach distinct from CNNs or Transformers, is designed to capture long-sequence data with the promise of linear scaling in sequence length, thereby attracting more attention to its mechanisms. 
%Building upon this foundation, the S5 layer~\citep{smithsimplified} integrates efficient parallel scanning and Multiple-Input Multiple-Output (MIMO) SSMs into the S4 framework. 
%In addition, H3~\citep{fuhungry} introduces a new SSM layer that significantly narrows the performance gap between SSMs and Transformer attention mechanisms in language modeling. 
The Gated State Space layer~\citep{mehta2023long} enhances S4 by introducing more gating units, improving training efficiency and achieving competitive performance.  Mamba~\citep{gu2023mamba,dao2024transformers} utilizes input-dependent parameters instead of the constant transition parameters used in standard SSMs, enabling more intricate and sequence-aware parameterization. This dynamic and selective approach efficiently interacts between sequential states, addressing drawbacks in capturing sequence context and reducing linear time complexity.
% In comparison with constant transition parameters in standard SSMs, Mamba utilizes input-dependent parameters, allowing for more intricate, sequence-aware parameterization. 
%This approach includes directly incorporating the parameters from the input sequence $\mathbf{x}$, thus facilitating a richer representation of the sequence context. 
It empirically surpasses Transformers performance~\citep{dosovitskiy2021an,touvron2021training} across various sizes on large-scale datasets.
% shi2024multiscale
Inspired by the merits of Mamba, several works~\citep{zhu2024vision,liu2024vmamba,li2024mamba} have proposed new models that integrate the SSM mechanism into vision tasks.
%, such as classification, segmentation, and image restoration. 
Vim~\citep{zhu2024vision} uses bi-directional Mamba blocks instead of self-attention mechanisms to compress visual representation. Similarly, VMamba~\citep{liu2024vmamba} introduces a well-designed selective scanning approach to arrange the position of patch images in both horizontal and vertical directions. %Both achieve superior performance compared to well-established vision transformers on various tasks, while providing efficient computation. 

% In this work, we draw inspiration from SSMs and explore the robustness of the visual state space model against backdoor attacks from various perspectives, \textit{e.g.}, the impact of visual model with or without SSM on such attacks.

% By adopting selective SSMs, Mamba models not only maintain linear scalability with respect to sequence length but also demonstrate strong performance in language modeling tasks. 

\subsection{Backdoor Attack and Defense}
% 簡短敘述攻擊，舉例
% 簡短敘述防禦，分為兩類，舉例
% , i.e., backdoor.

% Backdoor attacks~\citep{gu2019badnets,li2022backdoor} are a type of causative attack in the training process of DNNs, where a malicious attacker injects a specific pattern and alters corresponding target labels to contaminate a portion of the training data.

Backdoor attacks~\citep{gu2019badnets,li2022backdoor} are a type of causative attack in the training process of DNNs, building the relationship between a trigger pattern and altering target labels. The poisoned model classifies benign data accurately but predicts the targeted labels when the backdoor is activated. In addition, more intricate triggers~\citep{li2021invisible,nguyen2021wanet,qi2023revisiting} have been developed to enhance the stealthiness of backdoor attacks. That is, invisible backdoor attacks can be optimized using various techniques (\textit{e.g.}, steganography~\citep{li2020invisible}, latent space constraint~\citep{zhao2022defeat}).

% , light reflection~\citep{liu2020reflection}

% without altering the labels of backdoor samples~\citep{turner2019label}. 
% saha2020hidden
% Furthermore, some studies have implanted backdoors by enabling attackers to modify model parameters~\citep{zhang2022how} or insert targeted trojans~\citep{rakin2020tbt}, without relying on data poisoning.
% min2024uncovering, zhu2023enhancing
To confront advanced attacks, research~\citep{huang2022backdoor, zhu2023enhancing, zhu2024neural} on backdoor defense is continually evolving. According to different scenarios in real-world applications, most backdoor defenses can be partitioned into two categories: (1) \textit{In-training defense} trains a clean model based on the defense principle from a given poisoned dataset. For instance, DBD~\citep{huang2022backdoor} integrates supervised and self-supervised learning to disperse the poisoned samples in the feature space and subsequently employs the updated classifier to identify them based on confidence scores. (2) \textit{Post-training defense} attempts to mitigate backdoor effects from a given model with a small fraction of benign data. For example, FT-SAM~\citep{zhu2023enhancing} employs a sharpness-aware optimizer to minimize the loss function while reducing the sharpness of the loss surface around backdoor triggers, thereby enhancing model generalization. NPD~\citep{zhu2024neural} inserts a learnable transformation layer as an intermediate layer into the backdoored model and then purifies the model by solving a well-designed bi-level optimization problem.

% For instance, DBD~\citep{huang2022backdoor} decouples the training process into three steps. First, it trains a feature extractor on poisoned samples using self-supervised learning to prevent their clustering in the feature space. Second, it updates the classifier through supervised learning. Finally, the classifier assesses and identifies the poisoned samples according to confidence scores when learning the classifier.

% NAD~\citep{li2021neural} first fine-tunes a backdoored model with clean samples (\textit{i.e.}, $5\%$ of the training set) so that the fine-tuned model, acting as a teacher, can facilitate the backdoor removal of the original backdoored model through neural attention loss applied after each residual. 
% Fine-pruning (FP)~\citep{liu2018fine} aims to prune neurons in the backdoored model that have divergent activations between poisoned and clean samples. 
% ShapPruning, inspired by Shapley values, accurately identifies and prunes poisoned neurons using reversed triggers. In addition, ShapPruning introduces a threshold discard and $\epsilon$-greedy algorithm to accelerate the estimation of Shapley values in the model.

\subsection{Backdoor Robustness of Generic Vision Models}
% 說明許多關注在研究通用性模型，例如...
% ViT是一種現有文獻多半專注...，其中表示xxx ViT不如CNN穩健...
% 另一篇說明Visual SSM比ViT更有穩健性。
% 而本篇論文主要探討SSM對Backdoor attack的穩健性。

More recently, there have been efforts~\citep{qiu2023towards,yuan2023you,subramanya2024closer} aimed at improving the robustness of generic backbones against backdoor attacks through empirical analysis. For example, empirical studies on the robustness of ViTs and CNNs under different triggers~\citep{yuan2023you} reveal that the self-attention mechanism makes ViTs more vulnerable to patch-wise triggers than to image-blending triggers. Another study~\citep{doan2023defending} describes distinctive patch transformations on ViTs and introduces an effective defense to enhance their backdoor robustness. Furthermore, compared with SOTA architectures, Feed Forward Networks, such as ResMLP~\citep{touvron2022resmlp}, are more robust than CNNs and ViTs against backdoor attacks~\citep{subramanya2024closer}. In contrast, we analyze the comparative backdoor robustness of VSSMs, ViTs, and CNNs, with a particular focus on the susceptibility introduced by SSMs.
 
% Furthermore, they disclose how attention mechanisms affect backdoor robustness in ViTs, a distinction not observed in CNNs. 

% Unlike existing works, we explore the backdoor robustness of VSS model, especially the evaluation of SSM mechanisms.

% Yuan et al. (2023), Doan et al. (2023) prior work discusses and reveals Subramanya et al. (2024) explore through the lens of backdoor attacks. 

%% file: sec/3_pre.tex
\section{Preliminary}
\subsection{State Space Model}\label{pre:ssm}

Drawing inspiration from control theory, the State Space Model (SSM) represents a continuous system that maps a 1D sequence $x(t) \in \mathbb{R}$ to $y(t) \in \mathbb{R}$ through a hidden state $h(t) \in \mathbb{R}^\mathsf{N}$ as:
\begin{align}
\begin{split}
 h'(t) = \mathbf{A}h(t) + \mathbf{B}x(t), \quad y(t) = \mathbf{C}h(t),
\end{split}
\label{Eq: SSM-Hidden}
\end{align}
where $\mathbf{A} \in \mathbb{R}^{\mathsf{N}\times\mathsf{N}}$ denotes the evolution parameters, $\mathbf{B} \in \mathbb{R}^{\mathsf{N}\times\mathsf{1}}$ and $\mathbf{C} \in \mathbb{R}^{\mathsf{1}\times\mathsf{N}}$ denote the projection parameters.

To incorporate SSMs into deep models, S4~\citep{gu2022efficiently} defines the system in terms of ($\mathbf{A}, \mathbf{B}, \mathbf{C}$) and the sampling step size $\Delta$. In contrast, Selective SSM (S6)~\citep{gu2023mamba} treats parameters ($\mathbf{B}, \mathbf{C}, \Delta$) as functions of input, thereby enhancing computational efficiency and achieving better content-aware reasoning. The discrete version of Eq.~(\ref{Eq: SSM-Hidden}) can be written in the following form:
\begin{align} \label{eq:2}
\begin{split} h_t = \mathbf{\overline{A}}h_{t-1} + \mathbf{\overline{B}}x_t, \quad y_t = \mathbf{C}h_t, 
\end{split}
\end{align}
where the parameters $\overline{\mathbf{A}}$ and $\overline{\mathbf{B}}$ are derived by zero-order hold (ZOH), defined as: 
\begin{eqnarray} \overline{\mathbf{A}} = \exp(\Delta \mathbf{A}), \quad \overline{\mathbf{B}} = (\mathbf{\Delta} \mathbf{A})^{-1}(\exp(\Delta \mathbf{A}) - \mathbf{I}) \cdot \Delta \mathbf{B}.
\end{eqnarray}
Finally, the model computes the output $\mathbf{y}$ using a global convolution layer (denoted by $\circledast$) as $\mathbf{y} = \mathbf{x} \circledast \mathbf{\overline{K}}$, where the structured convolutional kernel $\overline{\mathbf{K}} \in \mathbb{R}^\mathsf{M}$ is defined as $\mathbf{\overline{K}} = (\mathbf{C}\mathbf{\overline{B}},\mathbf{C}\overline{\mathbf{A}\mathbf{B}}, ..., \mathbf{C}\mathbf{\overline{A}}^{\mathsf{M}-1}\mathbf{\overline{B}})$ and $\mathsf{M}$ is the length of the input sequence $\mathbf{x}$.

% Finally, the model computes the output through a global convolution layer (denoted by $\circledast$) as:  with $\mathbf{\overline{K}} = (\mathbf{C}\mathbf{\overline{B}},\mathbf{C}\overline{\mathbf{A}\mathbf{B}}, ..., \mathbf{C}\mathbf{\overline{A}}^{\mathsf{M}-1}\mathbf{\overline{B}})$, where $\mathsf{M}$ is the length of input sequence $\mathbf{x}$ and $\overline{\mathbf{K}} \in \mathbb{R}^\mathsf{M}$ is a structured convolutional kernel. 

% In this way, the model can use convolution to generate outputs across the sequence at the same time, improving computational efficiency and scalability.

\subsection{Formulation of VSSMs}
For a given VSSM, the image $t \in \mathbb{R}^{H \times W \times C}$ is transformed into flattened patches $x_p \in \mathbb{R}^{N \times (P^2 \times C)}$, where $H \times W$ represents the image size, $C$ denotes the number of channels, $N$ is the number of patches, $P$ is the size of image patches. 
%The patch matrix $x_p$ is then linearly projected to a vector of size $D$, and position embeddings $E_{pos} \in \mathbb{R}^{(N+1)\times D}$ are added. 
In this paper, we utilize Vim~\citep{zhu2024vision} as the VSSM, with the token sequence $\mathbf{T}_{0}$ defined as:
\begin{align}
\mathbf{T}_{0} &= [t_{cls};t_p^1\mathbf{M};t_p^2\mathbf{M};\cdots;t_p^{N}\mathbf{M}] + E_{pos},
\label{Eq: VSS-Hidden}
\end{align}
where $t^N_p$ is the $N$-th patch of image $t$, $\mathbf{M} \in \mathbb{R}^{(P^2 \cdot C)\times D}$ is the learnable projection matrix. The position embedding $E_{pos} \in \mathbb{R}^{(N+1)\times D}$ consists of $(N+1)$ vectors of dimension $D$, obtained by projecting the patch matrix $x_p$ into the $D$-dimensional space. Vim uses class token $t_{cls}$ to represent the whole patch sequence. The token sequence $\mathbf{T}_{\ell-1}$ is then fed into the $\ell$-th layer of the Vim encoder, and obtains the output $\mathbf{T}_{\ell}$. At last, the output class token $\mathbf{T}_{L}^{0}$ is normalized and fed into the multi-layer perceptron (MLP) head to obtain the final prediction $\mathtt{pred}$ as:
\begin{align}
\mathbf{T}_{\ell} &= \mathbf{Vim}(\mathbf{T}_{\mathtt{\ell}-1}) + \mathbf{T}_{\mathtt{\ell}-1},\\
\mathtt{pred} &= \mathbf{MLP}(\mathbf{Norm}(\mathbf{T}_{L}^0)),
\end{align}
where $L$ denotes the number of layers and $\mathbf{Norm}$ represents the normalization layer. Specifically, the Vim encoder first normalizes the input token sequence $\mathbf{T}_{\mathtt{\ell}-1}$ and then linearly projects it to $x$ and $z$ with dimension size $E$ (\textit{i.e.}, expanded state dimension). Next, a 1D convolution is applied to $x$, followed by linear projections that produce the parameters $\mathbf{B}$, $\mathbf{C}$, and $\Delta$, respectively. The parameter $\Delta$ is then used to transform $\overline{\mathbf{A}}$ and $\overline{\mathbf{B}}$ in Equation~\ref{eq:2}. Finally, the output $y$ is computed through Equation~\ref{eq:2}, gated by $z$, and added to obtain the output token sequence $\mathbf{T}_{\mathtt{\ell}}$.

%% file: sec/4_ours.tex
\section{The Proposed BadVim Framework}

% In this section, we first discuss the threat model for backdoor attacks on VSSMs. We also introduce the motivation of our attack and the formulation of attacks in VSSMs. Then, we explain how BadVim exploits the frequency sensitivity of SSM and demonstrate how our attack persists through state evolution in VSSMs.

\subsection{Threat Model} % 說明這篇使用的情境設定
\noindent \textbf{Adversary's capability:} A backdoor adversary aims to embed a backdoor function into the target model. We define the benign model as $\mathcal{F}$ and the backdoored model as $\hat{\mathcal{F}}$. The adversary's objective is to cause $\hat{\mathcal{F}}$ to misclassify all test samples from non-target classes as the target class $y_t$ when a predefined backdoor trigger $\mathcal{T}$ is present. Additionally, the adversary aims to maintain accuracy under benign samples by ensuring that $\hat{\mathcal{F}}$ correctly classifies benign test inputs, \textit{i.e}, those that do not contain $\mathcal{T}$.

\noindent \textbf{Adversary's knowledge:} We consider the adversary with minimal knowledge of the training pipeline, \textit{e.g.}, the training data or its distribution. Specifically, the adversary is unaware of the model architecture or parameters and cannot modify training settings, manipulate gradients, or alter the loss function. Therefore, in our threat model, backdoor attacks are conducted through data poisoning, where a subset of samples and their corresponding ground-truth labels are tampered with to embed backdoors into the VSSM during training.

% Considering that pre-trained models are commonly fine-tuned for various applications, we follow the typical threat model used in prior works~\citep{gu2019badnets,nguyen2021wanet,li2021invisible,qi2023revisiting}. 

\begin{figure}[!t]
    \centering
    \includegraphics[width=0.9\columnwidth]{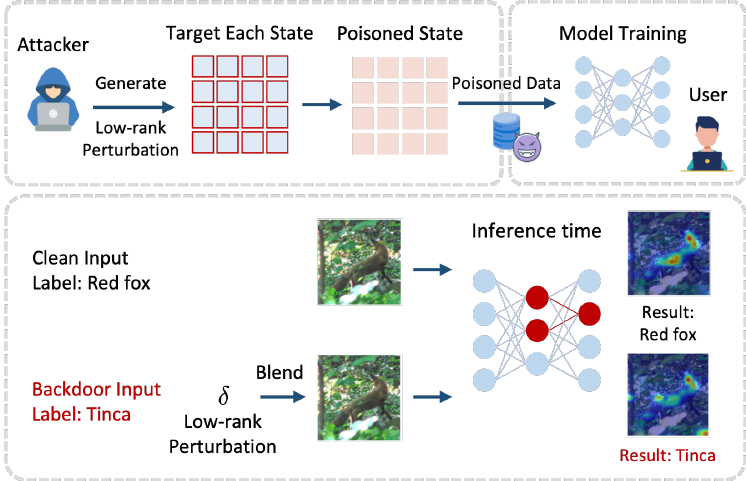}
    \vspace{1.0mm}
    \caption{Overview of BadVim framework. Poisoned samples are crafted by injecting low-rank perturbations $\delta$ into each patch state. An honest user downloads the poisoned data and trains a model, which malfunctions when a trigger is present during inference. Notably, the hidden attention matrices~\citep{ali2024hidden} of poisoned samples remain similar to those of clean samples.}
    \vspace{0.5cm}
    \label{fig:ours}
\end{figure}

% Figure 1. Framework of BadVim
% Heatmap of Visualization (Vim)
% \begin{figure*}[!ht]
%     \centering
%     \includegraphics[width=\textwidth]{fig/png/ours.png}
%     \vspace{0.1mm}
%     \caption{Overview of the VSS model and our proposed backdoor injection method. The left section (blue dashed box) illustrates the VSS block, including the SSM formulation and sequence scanning (\textit{e.g}, bidirectional scanning in Vim~\citep{zhu2024vision}), omitting normalization and shortcuts for simplicity. The right section (red dashed box) depicts our backdoor injection approach, where poisoned samples are generated by adding low-rank perturbations. Notably, the hidden attention matrices~\citep{ali2024hidden} of poisoned samples remain similar to those of clean samples.}
%     \vspace{0.5cm}
%     \label{fig:ours}
% \end{figure*}

\subsection{Intuition Behind our Backdoor Attack}
For simplicity, we outline the computational concept behind the SSM mechanism (\textit{e.g.}, S4~\citep{gu2022efficiently}, Mamba~\citep{gu2023mamba}), which applies to any VSSM utilizing low-rank matrix approximation~\citep{clarkson2017low} to reduce the time complexity of state updates. Due to space limitations, we present the key concept of Normal Plus Low-Rank (NPLR) here and defer the details of S4 to the full version.

% Section~\ref{abl:s4} of Supplementary Material.
% markovsky2012low
Recall that S4 represents the HiPPO matrix~\citep{gu2020hippo} as an NPLR decomposition, which parameterizes the state transition matrix as:
\begin{equation}\label{Eq: SSM-NPLR}
    \mathbf{A} = \mathbf{N} + \mathbf{P}\mathbf{Q}^{\intercal}, 
\end{equation} 
where $\mathbf{N}$ is a normal matrix (\textit{i.e.}, it satisfies $\mathbf{N}^{\intercal}\mathbf{N} =\mathbf{N}\mathbf{N}^{\intercal}$), and $\mathbf{P}$, $\mathbf{Q}$ are low-rank factors that restore interactions lost during diagonalization. To further interpret Equation~\ref{Eq: SSM-NPLR}, consider its relation to singular value decomposition (SVD). S4 and Mamba sets $\mathbf{N}$ as the diagonal matrix with negative elements, representing a special case of SVD in that the matrix contains only diagonal elements: $\mathbf{N} = \mathbf{U}\Lambda\mathbf{U}^{\intercal}$, where $\mathbf{U}$ is an orthogonal matrix and $\Lambda$ is a diagonal matrix of eigenvalues. The low-rank term $\mathbf{P}\mathbf{Q}^{\intercal}$ in Equation~\ref{Eq: SSM-NPLR} serves as an approximation and, in SVD form, contains only a few nonzero singular values, indicating a rank significantly lower than that of $\mathbf{A}$. The low-rank components $\mathbf{P}$ and $\mathbf{Q}^{\intercal}$ can be expressed as: 
\begin{align}
\mathbf{P}\mathbf{Q}^{\intercal} = \sum_{i=1}^{r}{\lambda_{i}\mathbf{u}_{i}\mathbf{v}_{i}^{\intercal}}, 
\label{Eq: SSM-LR}
\end{align} 
where $r$ is the rank of $\mathbf{P}\mathbf{Q}^{\intercal}$, and $\lambda_{i}$ and $\mathbf{u}_{i}$/$\mathbf{v}_{i}$ are the singular values and vectors corresponding to this low-rank structure. This makes SSM mechanisms more sensitive to low-rank perturbations in $\mathbf{A}$. However, in practice, VSSMs employs the discrete form of Equation~\ref{Eq: SSM-Hidden} for training. Moreover, an adversary cannot directly manipulate model parameters. Thus, we use data poisoning to indirectly perturb $\overline{\mathbf{A}}$ in VSSMs, as shown in Figure~\ref{fig:ours}.

% This makes it prone to learning triggers with low-rank structures.

% By adopting an NLPR representation, S4 and its successors (\textit{e.g} Mamba) inherit an implicit sensitivity to low-rank patterns.

% Essentially, the success rate of backdoor attacks depends on the ability of models to capture the correlation between the trigger and target class, so the question we face is “How to generate a trigger that can more effectively attract the attention of the model?”. We have gained insight from the investigation in Sec. 4 that ViTs are more vulnerable than CNNs under patch-wise triggers, so we intend to design a universal optimized patch-wise trigger that can fully focus the model’s attention on the area where it is located, so as to achieve a backdoor attack that is more stealthy, transferable, and less dependent on poisoning data.

% S4 considers matrix diagonalization on $\mathbf{A}$ for simplifies computation, rewriting the state transition matrix as $\mathbf{A} = \mathbf{V}\Lambda\mathbf{V}^{-1}$, where $\Lambda$ is a diagonal matrix containing the eigenvalues of $\mathbf{A}$ and $\mathbf{V}$ is the eigenvector matrix. Since diagonal matrices are easy to exponentiate and compute with, this transformation significantly reduces the computational complexity of state updates. However, the strict diagonalization limits expressivity because it does not introduce interactions beyond the eigenbasis.

\subsection{Our Low-Rank Perturbation Triggers}
Building on the properties of SSMs, we propose three trigger patterns that indirectly perturb $\overline{\mathbf{A}}$ in SSMs, as illustrated in Figure~\ref{fig:ours}. Each trigger pattern is carefully crafted to influence the model's internal state through periodic structures. Let $I(x,y)$ denote the image intensity at pixel location $(x,y)$, which forms the basis for embedding these patterns into the image.

\noindent \textbf{LRP-Sine.} The 2D sinusoidal wave trigger introduces a periodic structure, defined as
\begin{displaymath}
    I(x, y) = \sum_k[a_k sin(2\pi\omega_k x +\phi_k) + b_k sin(2\pi\omega_k y +\psi_k)].
\end{displaymath}
Here, $a_k$ and $b_k$ denote the amplitudes of the $k$-th frequency component along the $x$- and $y$-axes, respectively. $\omega_k$ and $\phi_k$/$\psi_k$ represent each direction's angular frequency and phase.

% The separability of sinusoidal components along both axes enables decomposition into rank-one elements, aligning with the low-rank formulation in SSMs, where dominant frequency components correspond to leading singular vectors.

\noindent \textbf{LRP-Checkerboard.} The checkerboard pattern can be seen as an alternating intensity structure, defined as $I(x, y) = (-1)^{(x+y)}$, which represents a periodic arrangement of high-contrast regions. It can be expressed as:
\begin{displaymath}
    I(x, y) = sign(sin(2\pi f_x x)sin(2\pi f_y y)).
\end{displaymath}
This pattern can be decomposed into a sum of rank-one components, similar to Equation~\ref{Eq: SSM-LR}.

\noindent \textbf{LRP-Stripes.} The stripe pattern exhibits intensity variations along a specific spatial direction, defined as $I(x,y) = \operatorname{sign}(\sin(2\pi f_x x))$. The orientation of the sinusoidal function determines the stripe direction, which can be extended to two dimensions as
\begin{displaymath}
    I(x, y) = sign(sin(2\pi f_x x)+sin(2\pi f_y y)),
\end{displaymath}
where $f_x$ and $f_y$ denote the spatial frequency along each axis.

% These triggers align with the low-rank decomposition of SSMs, where the term $\mathbf{P}\mathbf{Q}^{\intercal}$ in Eq.~(\ref{Eq: SSM-LR}) approximates the system’s dominant structure and, in SVD form, retains only a few dominant singular values. 

\begin{figure}[!t]
    \centering
    \includegraphics[width=0.95\columnwidth]{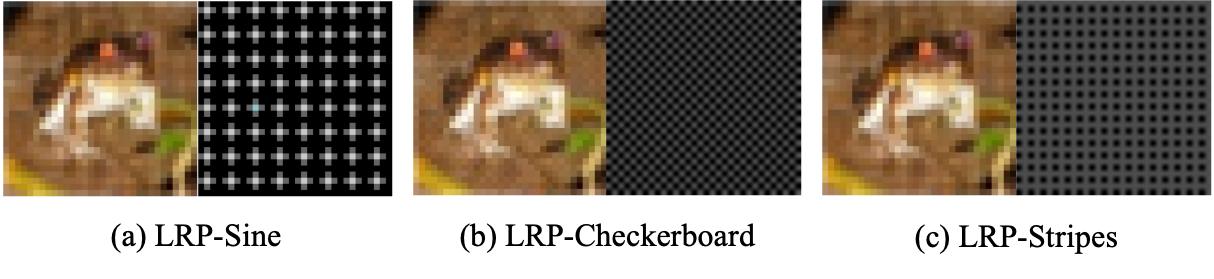}
    \caption{Illustration of the three strategies in BadVim. Here, the trigger is magnified 20 times for clarity.}
    \vspace{0.5cm}
    \label{fig:trigger}
\end{figure}

We assume that the adversary can arbitrarily choose any of these triggers for data poisoning. Each trigger is mapped to the pixel domain before embedding periodic patterns into clean images via \textit{image blending}. However, unlike ViTs, which apply global attention to the entire input simultaneously, VSSMs update their internal state iteratively, following different state-scanning orders (Figure~\ref{fig:illust}). This process can disrupt the trigger structure, reducing its effectiveness in VSSM. To ensure consistent backdoor implantation, we argue that \textit{the trigger must be embedded in each patch (state)}, regardless of the scanning trajectory (\textit{e.g.}, diagonal, Hilbert, or zigzag scans). Thus, we apply LRP at the patch level to influence every state in the sequence. As shown in Figure~\ref{fig:trigger}, we present three strategies applied to the same input, with the trigger magnified $20\times$ for clarity. Our main results use \textit{LRP-Checkerboard}. Due to space constraints, detailed algorithms with three patterns are provided in the full version.

% Section~\ref{abl:badvim} of Supplementary Material.

% Figure 2. Heatmap of Visualization (Vim)

% Similar to the previous cases, this structure decomposes into low-rank components, with dominant periodic elements corresponding to the leading singular vectors in \cref{Eq: SSM-LR}.

% This formulation highlights the periodic nature of the chessboard pattern, aligning it with Fourier basis functions.

% where $f_x$ and $f_y$ determine the spatial frequency along the $x$- and $y$-axes, respectively. This formulation introduces a two-dimensional periodic structure with alternating intensities, analogous to Fourier basis functions. The resulting pattern retains its structured periodicity, making it well-suited for analysis through Fourier decomposition and low-rank approximations, similar to those leveraged in SSMs.

% which describes a set of vertical or horizontal stripes, depending on the orientation of the sinusoidal function. More generally, the stripes pattern can be extended to exhibit periodicity along both spatial dimensions: 

\subsection{Why BadVim Works: Frequency Sensitivity in SSMs}
To further illustrate how BadVim influences SSM, we revisit the interpretation of the SSM mechanism as a global convolutional model, akin to CNNs, that exhibits similar frequency-selective behavior, as discussed in \citep{ravanelli2018speaker}. In CNNs, convolutional filters act as band-pass filters, selectively amplifying or attenuating specific frequency components. This allows CNNs to extract localized spectral patterns that reflect the structural characteristics of the input data. Similarly, SSMs capture certain frequencies through their learned kernel matrix $\mathbf{\overline{K}}$ (detailed in Section~\ref{pre:ssm}). The following equation further illustrates its connection to CNNs: 
\begin{align} 
S(t) = (\mathbf{I} \circledast \mathbf{W})(t), 
\end{align} 
where $\mathbf{W}$ denotes the trainable kernel of the CNN, $\mathbf{I}$ is the input sequence, and $S(t)$ is the output at time step $t$. When sinusoidal perturbations are introduced, the model’s band-pass characteristics selectively amplify those frequency components, causing it to increasingly attend to the perturbed frequency band and thereby enhancing the effectiveness of BadVim.

While SSMs share convolutional similarities with CNNs, recent studies~\citep{cirone2024theoretical} reveal that SSMs (\textit{e.g.}, Mamba) are limited in their ability to approximate arbitrary continuous input functions, leading to reduced expressivity. This constraint makes them prone to aligning with dominant input patterns--particularly periodic ones introduced by low-rank perturbations. As a result, SSMs tend to more aggressively amplify or suppress specific frequency components, reinforcing the effect of structured signals. Moreover, since Mamba updates its state based on input, the transition matrix $\overline{\mathbf{A}}$ often aligns with the perturbation frequency. This alignment propagates through the parameters $\overline{\mathbf{B}}$ and $\mathbf{C}$, leading to deviations from the model's expected behavior. Since $\overline{\mathbf{K}}$ encodes information from all these parameters, these cumulative effects shape the model's operation, consistent with the convolutional form.

% This constraint narrows the range of functions the model can approximate, making it prone to overfitting or alignment with dominant input patterns--especially periodic ones introduced by low-rank perturbations. 

Building on the above analysis, we observe that periodic perturbations introduced during training induce a frequency-dependent bias in the model's parameters. This makes the model more sensitive to backdoor triggers operating within the same frequency band. Furthermore, our analysis of the SSM's state dynamics reveals that such perturbations persist throughout the state transitions, which we refer to as perturbation persistence, as formalized below.

\begin{proposition}[Perturbation Persistence]\label{thm:impact}
    Consider the discrete state-space model: $h_t = \mathbf{\overline{A}}h_{t-1} + \mathbf{\overline{B}}x_t, \quad y_t = \mathbf{C}h_t$. Suppose that the input is corrupted by a low-rank perturbation $\delta$, \textit{i.e.}, $\hat{x}_t= x_t + \delta$. Let $\mathbf{\overline{A}'}$ denote the effective state transition matrix after perturbation. Then, the matrix $\mathbf{\overline{A}}$ has spectral radius $\rho(\mathbf{\overline{A}})>1$, the perturbation persists in the state evolution, leading to a spectral radius $\rho(\mathbf{\overline{A}'}) \geq 1$.
    % under suitable conditions on $\mathbf{\overline{A}}$, 
\end{proposition}    

Specifically, under the conditions on $\mathbf{\overline{A}'}$, this persistence leads to a spectral radius $\rho(\mathbf{\overline{A}'}) \geq 1$, indicating that the perturbation exerts a lasting influence on the model's behavior and amplifies the backdoor effect. This result supports the intuition that frequency-aligned perturbations persist through state transitions and accumulate in the hidden state over time, rather than dissipating. We provide a detailed proof in the full version of the paper.

%% file: sec/5_expt.tex
\section{Experiments}\label{expt}
\subsection{Evaluation Settings}
\noindent \textbf{Datasets and Models.}\quad Three common datasets, including CIFAR10~\citep{krizhevsky2009learning}, GTSRB~\citep{stallkamp2011german}, and ImageNet~\citep{deng2009imagenet}, were selected for the backdoor poisoning attack. For the VSSM, we adopted Vim-t~\citep{zhu2024vision}, the tiny-size variant of Vim, which employs bidirectional scanning, the simplest form of VSSM, while other variants use more complex scanning methods. Additionally, we compared robustness across three architectures, including ResNet18~\citep{he2016deep}, DeiT-t~\citep{touvron2021training}. Since pre-trained models are only available for ImageNet, we trained these models from scratch on CIFAR-10 and GTSRB. Due to space constraints, details on datasets, model architectures, and hyperparameters are provided in the full version.

% Section~\ref{abl:expt-set} of Supplementary Material.
% and GatedCNN-t~\citep{yu2025mambaout}.

% Table 2. Compare with ResNet18, DeiT, Vim (CIFAR10, GTSRB)
\input{tab/tab1-main-results}
\noindent \textbf{Attack Settings.}\quad 
We utilize a \textit{checkerboard} pattern as the backdoor trigger for our main results. By default, we set the perturbation frequency to $8$ and the image-blending ratio to $0.2$ for generating poisoning data. Our method was compared with the SOTA backdoor attacks (\textit{i.e.}, BadNets~\citep{gu2019badnets}, Blend~\citep{chen2017targeted}, SIG~\citep{barni2019new}, Refool~\citep{liu2020reflection},  TaCT~\citep{tang2021demon}, ISSBA~\citep{li2021invisible}, Dynamic~\citep{nguyen2020input}, WaNet~\citep{nguyen2021wanet}, and Adaptive-based~\citep{qi2023revisiting}), following the default configuration in Backdoor-ToolBox~\citep{qi2023revisiting} for fair comparison. The poisoning rate was fixed at $0.3\%$ for all attacks, except for ImageNet, where a higher rate of $1\%$ was used. Additionally, since Backdoor-ToolBox does not implement the clean-label attack (CLB)~\citep{turner2019label} on GTSRB, we excluded this attack from GTSRB experiments. Further details are provided in our full paper.

% Section~\ref{abl:expt-set} of Supplementary Material.
% our method, $5\%$ for WaNet, and below $3\%$ for the others. 

\noindent \textbf{Performance metrics.}\quad We used two common metrics to measure the effectiveness of backdoor attacks: Clean Accuracy (ACC) and Attack Success Rate (ASR). ACC measures the model's accuracy on benign test data without a backdoor trigger, while ASR evaluates the model's accuracy when tested on non-target class data with the specified trigger. Generally, a lower ASR coupled with a higher ACC indicates robustness to such attacks. 

\subsection{Empirical Results}\label{expt:result}
\noindent \textbf{Negligible impact on clean accuracy.}\quad Table~\ref{tab:main-results} presents a comprehensive comparison between the VSSM and the other two models. ``None'' represents the benign model without backdoor attacks. As can be seen, most models retain competitive ACC on clean samples when subjected to backdoor attacks. Specifically, our attacks on the Vim-t reduce ACC by less than $1$\%, demonstrating superior performance across three datasets compared to other attacks. Interestingly, we find that Dynamic attack achieves the second-best ACC in CIFAR and GTSRB. We speculate that this is due to its ability to adapt to varying input patterns, which may enhance robustness against perturbations.

Compared to ResNet18 and DeiT-t, Vim-t exhibits a lower average ACC on small-size datasets, \textit{e.g.}, $84.97$\% on CIFAR10, while it achieves the highest ACC on ImageNet (\textit{i.e.}, $76.76$\%). This indicates that VSSM benefits from its ability to model long-range dependencies through its state-space representation. Furthermore, we observe that our attacks maintain high ACC across all three models on ImageNet, indicating that the models generalize well to large-scale datasets, with minimal degradation in clean accuracy. 

% for GTSRB, we observe an increase in ACC in most attacks. This is because BadNets and WaNets require a higher poisoning rate ($10$\% and $5$\%, respectively) to be effective, yet their ASR remains relatively low ($48.49$\% and $35.46$\%). 

% In summary, these models preserve their original functionality on clean samples, with only a minor decline in accuracy under backdoor attacks.

% are implanted backdoors with a small decline in accuracy (around less than 1\% average accuracy decline on CIFAR10). 

% Similarly, Vim-t and MambaOut-t effectively maintain accuracy close to their original values (\textit{e.g.}, $85.35$\% and $82.46$\% for BadNets on CIFAR10, respectively).

% This discrepancy may be due to the VSS model's focus on contextual patch correlations via the SSM module, which is susceptible to backdoor attacks in terms of accuracy. 

\noindent \textbf{High attack success rate (ASR).}\quad In Table~\ref{tab:main-results}, we observe that Vim-t exhibits lower ASR for most invisible backdoors (\textit{e.g.}, $13.48$\% and $2.11$\% in Refool and WaNet on CIFAR10, respectively) compared to visible backdoors. Patch-based triggers, such as BadNets and TaCT, are less effective against Vim-t on small datasets due to the smaller patch size in VSSM, where the trigger is placed at a specific location, causing it to fragment or be disrupted when the image is divided into patches. Vim-t is particularly vulnerable to blending-based and input-aware (\textit{i.e.}, Dynamic) backdoors. This is because SSMs excel at establishing causal correlations between the patches and processing long-sequence inputs, facilitating the synthesis of global triggers. Furthermore, our attacks achieve high ASR across three datasets, \textit{e.g.}, $97.89$\% on ImageNet, demonstrating their effectiveness on VSSMs.

In comparison with different models, we show that the backdoor robustness of Vim-t is comparable to that of DeiT-t and superior to ResNet18. For instance, on CIFAR10, the average ASR for DeiT-t and Vim-t is $43.32$\% and $38.27$\%, respectively, while ResNet18 has an average ASR of $68.47$\%. Interestingly, our attack successfully injects backdoors into these models (\textit{i.e.}, at higher ASRs). These observations further validate that our attack could transfer to other model architectures. Moreover, this transferability extends across different model architectures, as evidenced by the VSSM's susceptibility to the attack. Additionally, these results emphasize the backdoor robustness of VSSM, suggesting that while state scanning in SSMs enhances ACC, it remains vulnerable to such attacks.

% which is state scanning in SSMs remains vulnerable to such attacks.

% Furthermore, we find that ISSBA, acting as a subtle perturbation similar to our attack, makes Vim-t especially susceptible on CIFAR-10 and GTSRB.

% Notably, MambaOut-t demonstrates the strong backdoor robustness among these models. Since the absence of SSM, the Gated CNN model demonstrates superior performance than the VSS model in most cases. This suggests that the SSM module, which processes inputs along a scanning trajectory, can inadvertently create a strong correlation between the target class and the trigger. As a consequence, the design of the VSS model should account for robustness against backdoor attacks to mitigate potential threats in real-world scenarios. 

\begin{table*}[!ht]
    \centering
    \vspace{0.1mm}
    \caption{Efficacy of model-agnostic defenses against our BadVim.}
    \vspace{0.1cm}
    % \vspace{-0.15in}
    \resizebox{0.6\textwidth}{!}{
    \setlength{\tabcolsep}{1mm}{
    \begin{tabular}{c|cc|cc|cc|cc|cc|cc}
    \toprule
    \multicolumn{1}{c|}{\multirow{2}{*}{Dataset}} &\multicolumn{2}{|c|}{No Defense} & \multicolumn{2}{|c}{PatchDrop} & \multicolumn{2}{|c}{PatchShuffle} & \multicolumn{2}{|c}{FT} & \multicolumn{2}{|c}{FP} & \multicolumn{2}{|c}{FT-SAM} \\ \addlinespace[0.2em]
    \cline{2-13} \addlinespace[0.2em]
     & ACC & ASR & ACC & ASR & ACC & ASR & ACC & ASR & ACC & ASR & ACC & ASR  \\ \midrule
    CIFAR-10 & 85.60 & 100 & 81.83 & 100 & 69.26 & 100 & 88.86 & 99.97 & 88.71 & 99.97 & 87.46 & 99.94 \\ 
    GTSRB & 92.42 & 99.90 & 91.95 & 100 & 52.18 & 99.20 & 98.93 & 99.91 & 98.84 & 99.94 & 99.33 & 97.15 \\ 
    ImageNet & 77.11 & 97.89 & 76.44 & 94.06 & 65.93 & 93.14 & 75.48 & 69.83 & 68.46 & 52.96 & 67.80 & 24.32 \\  \bottomrule
    \end{tabular}}}\label{tab:vit-bd-defense}
\end{table*}

\subsection{Resistance to Backdoor Defenses}
% Table 3. and Table 4. Defense & Detection
% \input{ICCV2025-Author-Kit/tab/tab3}
% \input{ICCV2025-Author-Kit/tab/tab4}

% Table 3 and Table 4
Due to space limitations, we examine the performance of five existing backdoor defenses and two input-level detection methods against our attack here. We defer a detailed discussion to our full paper.
% Section~\ref{abl:expt-set} of Supplementary Material.

\noindent \textbf{(1.)} PatchDrop~\citep{naseer2021intriguing} and PatchShuffle~\citep{kang2017patchshuffle}, two patch-processing defense methods for ViTs~\citep{doan2023defending}, do not perform as well to eliminate BadVim. As shown in Table~\ref{tab:vit-bd-defense}, the VSSM retains a high ASR (\textit{e.g.}, $94.06\%$ on PatchDrop) under our trigger pattern, while ACC drops significantly (\textit{e.g.}, $65.93\%$ on PatchShuffle) due to the disruption of spatial information.

\noindent \textbf{(2.)} Fine-tuning (FT) and fine-pruning (FP)~\citep{liu2018fine} exhibit high ASR in our attack, even with extra clean data (\textit{i.e.}, $5$\% of the training set). From Table~\ref{tab:vit-bd-defense}, we find that the backdoor behavior is pronounced in small-scale datasets (\textit{e.g.}, $99.97\%$ ASR on CIFAR-10), and although reduced on ImageNet (to $28.06\%$ for FT and $44.93\%$ for FP), this still does not eradicate backdoors.
% --highlighting the robustness of BadVim on VSSMs.

% However, our attack remains a significant concern, as ASR for both methods consistently exceeds $50\%$.

\noindent \textbf{(3.)} Fine-tuning with sharpness-aware minimization (FT-SAM)~\citep{zhu2023enhancing} is the most effective defense among those we evaluated. On ImageNet, it considerably reduces ASR to $24.32\%$, but performs poorly on CIFAR-10 and GTSRB (\textit{e.g.}, $97.15\%$ ASR). Thus, FT-SAM is ineffective in fully cleansing the backdoored model.

\noindent \textbf{(4.)} Strip~\citep{gao2019strip} and cognitive distillation (CD)~\citep{huang2023distilling}, which focus on input-level backdoor detection, fail to achieve higher TPR and lower FPR on both datasets; the model would still retain backdoors after training, as shown in Table~\ref{tab:vit-bd-detect}. This is because BadVim introduces only an imperceptible perturbation in the spatial domain, making it difficult to identify the backdoor samples.

\begin{table}[!h]
    \centering
    \vspace{-0.1mm}
    \caption{Efficacy of input-level backdoor detections against BadVim.}
    \vspace{0.1cm}
    % \vspace{-0.15in}
    \resizebox{0.7\columnwidth}{!}{
    \setlength{\tabcolsep}{1mm}{
    \begin{tabular}{c|ccc|ccc}
    \toprule
    \multicolumn{1}{c|}{\multirow{2}{*}{Dataset}} &\multicolumn{3}{|c|}{Strip} & \multicolumn{3}{|c}{CD}  \\ \addlinespace[0.2em]
    \cline{2-7} \addlinespace[0.3em]
     & TPR & FPR & AUC & TPR & FPR & AUC  \\ \midrule
    CIFAR10 & 33.20 & 9.92 & 0.6766 & 80.39 & 15.73 & 0.8233 \\ 
    GTSRB & 1.71 & 9.86 & 0.3578 & 19.47 & 13.40 & 0.6122 \\  \bottomrule
    \end{tabular}}} \label{tab:vit-bd-detect}
\end{table}
\vspace{-0.15in}

\subsection{Interpretability of VSSM via Heatmaps}

\begin{figure}[!ht]
    \centering
    \includegraphics[width=0.8\columnwidth]{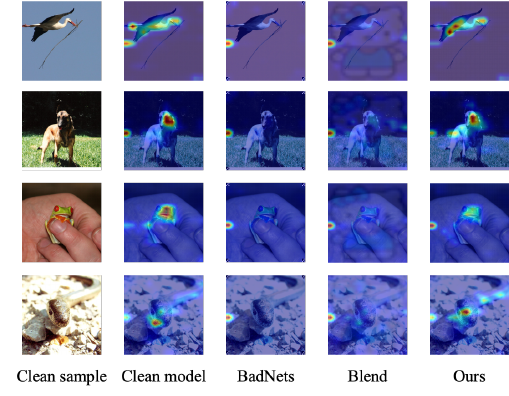}
    \vspace{0.1mm}
    \caption{Visualization of VSSM under implicit attention matrices~\citep{ali2024hidden} on ImageNet. We evaluate clean samples using the clean model (second column) and poisoned samples using the backdoored model (third to fifth columns), respectively.}
    % \vspace{0.25in}
    \label{fig:heatmap}
\end{figure}

To better understand how SSMs in the VSSM extract image representations, we leverage implicit attention matrices~\citep{ali2024hidden} within the SSM layer to analyze and interpret the behavior of backdoored models. As shown in Figure~\ref{fig:heatmap}, we observe that patch-based methods, such as BadNets, do not shift attention from the object to the trigger; instead, attention is diminished on the object and shifts to edges. This suggests that the backdoor in VSSM arises from the dirty label rather than the trigger pattern. In addition, we find that this phenomenon also occurs in Blend, which can be easily distinguished from benign samples due to its distinct behavior. In contrast, the heat maps generated from our attacks are similar to those of the clean and backdoor images. This is because the image is generated by adding an extremely small perturbation to the clean image; thus, the difference in the hidden attention, where the heat maps are generated, is also minimal. Consequently, existing defense methods~\citep{subramanya2024closer} that detect backdoors based on trigger-induced attention shifts in heatmaps fail to distinguish our backdoor samples from clean ones, rendering them ineffective against our attack.

% Figure 3. Heatmap of Visualization (Vim)

% To further validate our hypothesis, we present a simple yet effective trigger pattern that recurs in each patch. This pattern can be constructed using various frequency- or pixel-based techniques. Importantly, embedding the trigger pattern in each patch enhances resistance to perturbations. An intuitive method is to use a static trigger, such as a single pixel, placed repeatedly across the entire image. In this way, this pattern is consistently adjacent across the image, unaffected by spatial or geometric transformations. As illustrated in Fig.~\ref{fig:trigger}, our experiment set the distance between single pixels to $1$ and their intensity to $30$. From Table~\ref{tab:imnet-ssm-defense}, we find that our proposed trigger pattern maintains a high ASR (\textit{i.e.}, $94.06$\% on PatchDrop) with a $2$\% poisoning rate, even under these conditions. This observation supports our hypothesis that, despite patch processing, the VSS model is still susceptible to these remaining triggers.

%  In other words, the SSM module of the VSS model plays a pivotal role in fitting the data, making the VSS model more susceptible to backdoor attacks (\textit{i.e.}, patch-based trigger and the blending-based trigger) than the model that only stacks the Gated CNN blocks. This prompts further analysis of the impact of the absence of SSM modules on backdoor attacks in subsequent sections.

\subsection{Robustness Analysis: Architecture and Model Scale}
% Table 5. Compare with Vim, GatedCNNs (CIFAR10, GTSRB, ImageNet)

\noindent \textbf{Evaluation of BadVim on Different VSSMs.}\quad We evaluate the effectiveness of BadVim on another VSSM, VMamba~\citep{liu2024vmamba}, which differs from Vim in its scanning trajectory and architectural modules. As shown in Table~\ref{tab:imnet-vim}, VMamba-t achieves a notably higher ASR under our attack (\textit{i.e.}, $99.98$\%) than Vim-t, while maintaining high ACC. Compared to Vim-t, VMamba-t demonstrates stronger robustness against BadNets and Blend, achieving lower ASR values (\textit{i.e.}, $41.92$\% and $26.16$\%). These findings suggest that BadVim effectively exploits the sequential state processing patterns and inductive biases inherent to these models. Thus, both VSSMs are susceptible to BadVim, revealing a shared vulnerability within the model.

\input{tab/tab4-imnet-vim}
% \vspace{-0.15in}
% to assess its robustness.

\noindent \textbf{VSSM vs. Gated CNN.}\quad In Table~\ref{tab:imnet-ssm}, we compare the backdoor robustness of VSSMs and Gated CNNs, which differ only in the use of the SSM mechanism. We fine-tuned Vim-t~\citep{zhu2024vision} and GatedCNN-t~\citep{yu2025mambaout} for ten epochs. Vim-t shows high ASR under BadNets and Blend (up to $99.16$\%) but with a moderate drop in ACC (down to $75.06$\%). In contrast, GatedCNN-t achieves lower ASR (\textit{e.g.}, $35.30$\%) while retaining similar ACC (\textit{e.g.}, $78.51$\%), suggesting stronger robustness due to its ability to capture global patterns and semantic features across entire images through well-designed convolutional layers.

\input{tab/tab5-imnet-ssm}
% \vspace{-0.15in}

Our method maintains competitive ASR on both architectures (\textit{e.g.}, $97.89$\% on Vim-t and $98.32$\% on GatedCNN-t) with minimal ACC degradation. This is attributed to VSSM's emphasis on local features, as it processes images patch (state) by patch (state) to model long-range dependencies. These results demonstrate that BadVim, based on the low-rank perturbation, delivers consistent effectiveness across architectures. Detailed analysis is provided in our full version.

% Section~\ref{abl:insight} of the Supplementary Material.

% In Table~\ref{tab:imnet-ssm}, we compare the backdoor robustness of VSSMs and Gated CNNs, where the only difference between the two models is the inclusion of the SSM mechanism. Specifically, we fine-tuned Vim-t~\citep{zhu2024vision} and GatedCNN-t~\citep{yu2025mambaout} for ten epochs. We observe that Vim achieves a higher ASR (\textit{e.g}, up to $99.16$\%) but shows a slight decline in ACC (\textit{e.g}, down to $75.06$\%) under BadNets and Blend. In comparison, our attack maintains ACC close to the clean model with competitive ASR (\textit{e.g}, $97.89$\%). This is attributed to VSSM's emphasis on local features, as it processes images patch (state) by patch (state) to model long-range dependencies. In contrast, GatedCNN-t demonstrates stronger robustness, achieving a much lower ASR (\textit{e.g}, 35.30\%) yet retaining competitive ACC (\textit{e.g}, $78.51$\%). This suggests that Gated CNNs capture global patterns and semantic features across entire images through well-designed convolutional layers. Furthermore, we also find that our attack remains effective on GatedCNN-t, achieving a higher ASR (\textit{i.e}, $98.32$\%) with only a slight drop in ACC (\textit{i.e}, $78.68$\%). These results demonstrate that BadVim, based on the low-rank perturbation, delivers consistent backdoor performance across diverse models. Additional analysis is provided in Section~\ref{abl:insight} of the Supplementary Material.
\subsection{Ablation Studies}
\noindent \textbf{Impact of Model Size on Backdoor Robustness.}\quad In Table~\ref{tab:imnet-ssm-size}, we analyze the influence of different model sizes on the VSSM against backdoor robustness. We evaluated three pre-trained models from Vim, where our attack consistently achieves a high ASR, reaching $97.89$\% on Vim-t, while maintaining high accuracy (\textit{i.e.}, $77.11$\%), compared to BadNets and Blend. As the model size increases, Vim-s shows a slight decrease in ASR, reaching $82.35$\% for BadNets and $82.76$\% for Blend.  Vim-b reduces the impact of backdoor trigger features by distributing the learned information across more dimensions. This makes it harder for the attacker to exploit the specific trigger to induce backdoor behavior in the target model, improving its robustness against backdoor attacks.

% In addition, we observe that ACC on the Vim-s model drops less compared to the Vim-t model. This suggests that an increase in model size enhances backdoor robustness.

\input{tab/tab6-imnet-size}
% \vspace{-0.2in}

\begin{figure}[!ht]
    \centering
    \begin{minipage}[t]{0.48\columnwidth}
        \centering
        \includegraphics[width=0.85\linewidth]{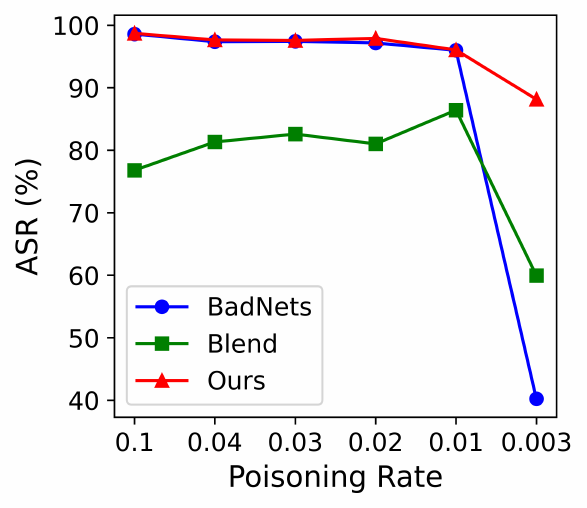}
        \caption{Effect of poisoning rate.}
        \label{fig:poison_rate}
    \end{minipage}
    \hfill
    \begin{minipage}[t]{0.48\columnwidth}
        \centering
        \includegraphics[width=0.85\linewidth]{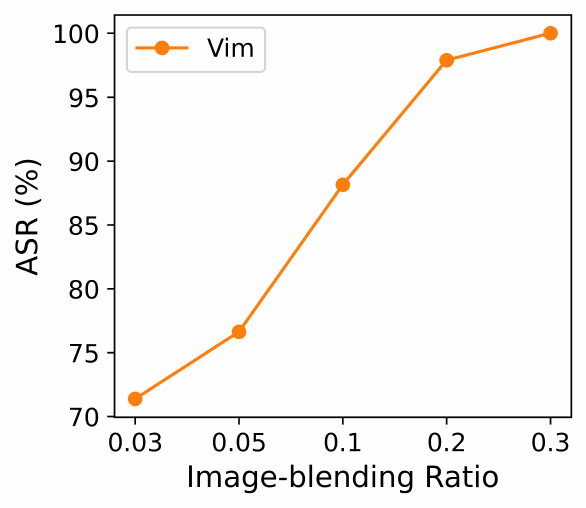}
        \caption{Effect of blending rate.}
        \label{fig:blend_rate}
    \end{minipage}
\end{figure}

\noindent \textbf{Effect of Poisoning Rate on ASR.}\quad We explore the impact of different poisoning rates on Vim-t by fine-tuning the pre-trained model. As shown in Figure~\ref{fig:poison_rate}, both BadNets and Blend achieve high ASR at poisoning rates as low as $1$\%.  However, at a poisoning rate of $0.3$\%, ASR drops significantly for both attacks, \textit{i.e.}, $40.22$\% for BadNets and $75.92$\% for Blend, indicating that the backdoor attack becomes less effective. In contrast, our proposed trigger maintains a high ASR of $88.14$\% even at an extremely low poisoning rate of $0.3$\%, showcasing its effectiveness against the VSSM.

\vspace{0.25in}

\noindent \textbf{Effect of Blending Rate on ASR.}\quad We examine the effect of varying blending ratios on Vim-t by fine-tuning the pre-trained model. As shown in Figure~\ref{fig:blend_rate}, ASR increases with higher blending ratios. We select a fixed blending ratio of $0.2$ to balance attack effectiveness and trigger stealthiness. Importantly, even at a low blending ratio of $0.03$, the ASR still exceeds $70\%$. This indicates that even slight perturbations are sufficient for the model to memorize the pattern, demonstrating the efficacy of BadVim against VSSMs.

% \vspace{0.25in}

% Table 1. Compare with DeiT, Vim (ImageNet)

%% file: tab/tab1-main-results.tex
\begin{table*}[!ht]
\centering
% \vspace{-0.1in}
\caption{The clean accuracy (ACC \%) and the attack success rate (ASR \%) of three models against backdoor attacks across three datasets, including CIFAR-10, GTSRB and ImageNet. The best results are bolded. The second-best results are underlined. Note that model names marked with ``t'' denote tiny size.}
% \vspace{-0.15in}
\vspace{0.1cm}
\setlength{\tabcolsep}{3mm}{
\resizebox{1.2\columnwidth}{!}{
\begin{tabular}{c|c|cc|cc|cc}
\toprule[0.68pt] \addlinespace[1.5pt]
\multicolumn{1}{c|}{\multirow{2}{*}{Dataset}} & \multirow{2}{*}{Attack} & \multicolumn{2}{c|}{ResNet18} & \multicolumn{2}{c|}{DeiT-t}  & \multicolumn{2}{c}{Vim-t}\\ \cline{3-8} \addlinespace[0.1em]
%\cmidrule{3-10}
& & \multicolumn{1}{c}{ACC} & \multicolumn{1}{c|}{ASR} & \multicolumn{1}{c}{ACC} & \multicolumn{1}{c|}{ASR} & \multicolumn{1}{c}{ACC} & \multicolumn{1}{c}{ASR} \\ \hline \addlinespace[0.1em]
\multicolumn{1}{c|}{\multirow{14}{*}{CIFAR-10}}  & \multicolumn{1}{c|}{None}& 94.90 & - & 90.95 &  -  & 86.19 & - \\ \cline{2-8} \addlinespace[0.1em]
\multicolumn{1}{c|}{}& BadNets & 94.08 & \textbf{100} & \underline{90.60} & 40.83 & 85.02 & 2.65 \\
\multicolumn{1}{c|}{}& Blend & \textbf{94.21} & 92.45 & 90.36 & 96.52 & 85.23 & 88.01 \\
\multicolumn{1}{c|}{}& SIG & \underline{94.11} & 57.68 & 90.42 & 68.26 & 85.43 & 54.27 \\
\multicolumn{1}{c|}{}& Refool & 93.17 & 16.43 & 90.51 & 4.17 & 85.52 & 13.48 \\
\multicolumn{1}{c|}{}& TaCT & 93.93 & 99.37 & 90.02 & 20.22 & 85.55 & 37.14 \\
\multicolumn{1}{c|}{}& CLB & 93.95 & \underline{99.90} & \textbf{90.63} & 54.19 & 83.47 & 13.27 \\
\multicolumn{1}{c|}{}& Dynamic & 93.91 & 98.81 & 90.53 & \underline{99.05} & \underline{85.58} & \underline{89.75} \\
\multicolumn{1}{c|}{}& ISSBA & 93.90 & 0.06 & 90.53 & 1.12 & 83.08 & 1.97 \\
\multicolumn{1}{c|}{}& WaNet & 93.23 & 1.62 & 89.98 & 1.39 & 84.73 & 2.11 \\
\multicolumn{1}{c|}{}& Adv-Patch & 93.73 & 99.12 & 90.28 & 3.00 & 85.07 & 7.89 \\
\multicolumn{1}{c|}{}& Adv-Blend & 93.76 & 51.31 & 90.33 & 38.12 & 85.06 & 56.70 \\
\multicolumn{1}{c|}{}& Ours & 93.53 & \textbf{100} & \textbf{90.63} & \textbf{100} & \textbf{85.61} & \textbf{100} \\ \cline{2-8} \addlinespace[0.1em]
& Average & 93.79 & 68.47 & 90.44 & 43.32 & 84.97 & 38.27 \\ \hline \addlinespace[0.1em]

\multicolumn{1}{c|}{\multirow{13}{*}{GTSRB}}  & \multicolumn{1}{c|}{None}& 97.22 & - & 94.10 & - & 91.27 & - \\ \cline{2-8} \addlinespace[1.5pt]
\multicolumn{1}{c|}{}& BadNets & \textbf{97.19} & 1.79 & 92.91 & 77.25 & 92.09 & 1.02 \\
\multicolumn{1}{c|}{}& Blend & 96.79 & \underline{92.24} & \underline{93.17} & 96.74 & 92.18 & \underline{93.70}  \\
\multicolumn{1}{c|}{}& SIG & 96.97 & 44.17 & 92.75 & 45.40 & 92.07 & 43.73  \\
\multicolumn{1}{c|}{}& Refool & 96.71 & 32.56 & 93.09 & 26.83 & 91.96 & 28.15  \\
\multicolumn{1}{c|}{}& TaCT & 97.08 & \textbf{100} & 93.04 & 8.32 & 92.38 & 8.24 \\ 
\multicolumn{1}{c|}{}& Dynamic & 97.00 & \textbf{100} & 92.79 & \underline{99.55} & 92.50 & 60.28 \\
\multicolumn{1}{c|}{}& ISSBA & 97.03 & 0.11 & 92.81 & 2.21 & \underline{92.71} & 3.86 \\
\multicolumn{1}{c|}{}& WaNet & 96.91 & 0.18 & 92.49 & 0.44 & 92.31 & 1.08 \\
\multicolumn{1}{c|}{}& Adv-Patch & \underline{97.14} & 21.73 & 92.75 & 1.63 & 92.23 & 4.75 \\
\multicolumn{1}{c|}{}& Adv-Blend & 96.66 & 78.02 & 92.89 & 74.86 & 92.45 & 81.84 \\
\multicolumn{1}{c|}{}& Ours & 96.72 & \textbf{100} & \textbf{93.51} & \textbf{100} & \textbf{93.64} & \textbf{100} \\ \cline{2-8} \addlinespace[0.1em]
& Average & 96.92 & 52.26 & 92.92 & 48.56 & 92.35 & 38.06 \\ \hline \addlinespace[0.1em]

\multicolumn{1}{c|}{\multirow{5}{*}{ImageNet}}  & \multicolumn{1}{c|}{None}& 69.75 & - & 70.07 & - & 77.29 & - \\ \cline{2-8} \addlinespace[0.1em]
\multicolumn{1}{c|}{}& BadNets & 69.21 & 75.86 & \underline{69.79} & \underline{95.54} & \underline{76.75} & \underline{97.16} \\ %Rate=0.02
\multicolumn{1}{c|}{}& Blend & \underline{69.24} & \underline{98.58} & 68.62 & 92.23 & 76.43 & 86.41 \\ %Rate=0.01
\multicolumn{1}{c|}{}& Ours & \textbf{69.67} & \textbf{99.24} & \textbf{69.85} & \textbf{99.37} & \textbf{77.11} & \textbf{97.89} \\ %Rate=0.02
\cline{2-8} \addlinespace[0.1em]
\multicolumn{1}{c|}{}& Average & 69.37 & 91.26 & 69.42 & 95.71 & 76.76 & 93.82 \\ \addlinespace[-0.2em]
\bottomrule[0.68pt]
\end{tabular}}} \label{tab:main-results}
\end{table*}
\vspace{0.1cm}

%% file: tab/tab4-imnet-vim.tex
\begin{table}[!ht]
\begin{center}
% \vspace{-0.1in}
\caption{Comparison between Vim and VMamba under Backdoor Attacks on ImageNet.}
\vspace{0.1cm}
% \vspace{-0.15in}
\setlength{\tabcolsep}{1mm}
\resizebox{0.75\columnwidth}{!}{%
\begin{tabular}{c|c|cc|cc|cc}
\toprule
\multicolumn{2}{c|}{Attacks Type} & \multicolumn{2}{|c|}{BadNets} & \multicolumn{2}{|c}{Blend} & \multicolumn{2}{|c}{Ours} \\ \addlinespace[0.2em]
\hline \addlinespace[0.2em]
Model & ACC & ACC & ASR & ACC & ASR & ACC & ASR \\ \addlinespace[0.2em]
\hline \addlinespace[0.2em]
Vim-t & 77.29 & \underline{76.75} & \underline{97.16} & 76.43 & 86.41 & \textbf{77.11} & \textbf{97.89}\\ 
VMamba-t & 81.84 & 81.51 & \underline{41.92} & \underline{81.55} & 26.16 & \textbf{81.88} & \textbf{99.98}\\
\bottomrule
\end{tabular}}\label{tab:imnet-vim}
\end{center}
\end{table}
% 75.06 & \textbf{99.16}

%% file: tab/tab5-imnet-ssm.tex
\begin{table}[!ht]
\begin{center}
\vspace{0.1mm}
\caption{Comparison of VSSMs and Gated CNNs under Backdoor Attacks on ImageNet.}
% \vspace{-0.15in}
\vspace{0.1cm}
\setlength{\tabcolsep}{1mm}
\resizebox{0.75\columnwidth}{!}{%
\begin{tabular}{c|c|cc|cc|cc}
\toprule
\multicolumn{2}{c|}{Attacks Type} & \multicolumn{2}{|c|}{BadNets} & \multicolumn{2}{|c}{Blend} & \multicolumn{2}{|c}{Ours} \\ \addlinespace[0.2em]
\hline \addlinespace[0.2em]
Model & ACC & ACC & ASR & ACC & ASR & ACC & ASR \\ \addlinespace[0.2em]
\hline \addlinespace[0.2em]
Vim-t & 77.29 & \underline{76.75} & \underline{97.16} & 76.43 & 86.41 & \textbf{77.11} & \textbf{97.89}\\
GatedCNN-t & 78.77 & 78.51 & 35.30 & \textbf{78.71} & \underline{73.34}  & \underline{78.68} & \textbf{98.32}\\
\bottomrule
\end{tabular}}\label{tab:imnet-ssm}
\end{center}
\end{table}

%% file: tab/tab6-imnet-size.tex
\begin{table}[h]
\begin{center}
\vspace{0.1mm}
\caption{Backdoor robustness evaluation on different sizes of VSSM with ImageNet. Model names marked with ``t'' denote tiny size, ``s'' small size, and ``b'' base size.}
% \vspace{-0.15in}
\vspace{0.1cm}
\setlength{\tabcolsep}{1mm}{
\resizebox{0.7\columnwidth}{!}{%
\begin{tabular}{c|c|cc|cc|cc}
\toprule
\multicolumn{2}{c|}{Attack} &\multicolumn{2}{|c|}{BadNets} & \multicolumn{2}{|c}{Blend} & \multicolumn{2}{|c}{Ours} \\ 
\midrule
Model & ACC & ACC & ASR & ACC & ASR & ACC & ASR \\
\midrule
Vim-t & 77.29 & \underline{76.75} & \underline{97.16} & 76.43 & 86.41 & \textbf{77.11} & \textbf{97.89}\\ 
Vim-s & 81.08 & 80.98 & \underline{82.35} & \textbf{81.10} & 82.76 & \underline{81.02} & \textbf{83.50} \\ 
Vim-b & 80.13 & 79.51 & \underline{72.92} & \underline{79.70} & 28.50 & \textbf{79.84} & \textbf{74.67} \\ 
\bottomrule
\end{tabular}}}\label{tab:imnet-ssm-size}
\end{center}
\end{table}

%% file: sec/6_discuss.tex
\section{Discussion and limitations}
\subsection{Backdoor Robustness: VSSMs vs. ViTs}
Recent studies ({\em e.g.}, \citep{han2024demystify}) suggest that the VSSM block can be viewed as a variant of linear attention, incorporating an input gate and a forget gate (\textit{i.e.}, $\overline{\mathbf{A}}$ in Equation~\ref{eq:2}). While these modifications improve task performance, the recurrent computation of the forget gate may be suboptimal for vision models. Moreover, as shown in Section~\ref{expt:result}, linear attention-like mechanisms do not enhance backdoor robustness and are more vulnerable to certain triggers (\textit{e.g}, patch-based and image-blend). Given structural similarities between ViTs and VSSMs---particularly their reliance on patch tokenization and linear projections---we conclude that their backdoor robustness is likely comparable as well.
% enhance performance across various tasks

%\subsection{Adversarial Robustness versus Backdoor Vulnerability}
% 引用adv & backdoor tradeoff 和原本的understanding那篇說明
\subsection{Adversarial vs. Backdoor Robustness in VSSMs}
Following the previous work~\citep{weng2020trade} that highlights a trade-off between adversarial and backdoor robustness, such a similar phenomenon is also observed in the VSSM. Specifically, the study~\citep{du2024understanding} indicates that the VSSM is more robust to adversarial attacks than ViTs due to the challenging gradient estimation. The trade-off between model size and robustness becomes more apparent as model size increases. In contrast, our findings reveal that \textit{the backdoor robustness of the VSSM is comparable to that of ViTs but slightly lower than that of GatedCNNs without SSM modules.} These results suggest that future research should consider backdoor robustness when developing novel architectures or defenses to mitigate potential threats.

% As discussed in previous sections, the SSM module within the VSS model demonstrates a strong learning capability in capturing feature information.

% As revealed in recent studies~\citep{han2024demystify}, Mamba can be interpreted as a variation of linear attention with several differences. These modifications contribute to its improvements in various performance metrics but do not alter its basis in terms of model architectures.

% Previous studies have shown that patch-based backdoor attacks often exploit the attention-based model. 

% Since Mamba and linear attention operate using linear projections without the non-linear softmax function, the linearity results in potentially higher attack success rates compare to conventional softmax-based attention.

% Given that both techniques share common properties, we can assume that their backdoor robustness should align closely as well.

\subsection{Limitations}
% We study the robustness of the SSM mechanism in VSSMs against backdoor attacks. However, we cannot delve into the impact of SSM parameters on backdoors, as tracking and accessing these parameters during training remains challenging. Our trigger design is based on patch sensitivity analysis, which may not be optimal. Lastly, while we focus on image classification, VSSMs are also used in other domains (\textit{e.g.}, NLP), which we plan to explore in future work.
We discuss the robustness of SSM mechanism in the VSSMs against backdoor attacks in this work. However, we cannot delve into the impact of SSM parameters on backdoors, as tracking and accessing these parameters during training remains challenging. Moreover, our trigger pattern for VSSM is based on patch sensitivity analysis, which may not necessarily be optimal. Finally, while our focus has been on the SSM mechanism of VSSM in image classification tasks, this mechanism has also been applied to other domains, such as natural language processing. We consider extending the research to these domains as part of our ongoing work.

%% file: sec/7_conclusion.tex
\section{Conclusions}

In this paper, we present a systematic study of the vulnerability of VSSMs to backdoor attacks, which exploit the state-space representation inherent in SSMs. Building on this insight, we propose BadVim, a low-rank perturbation approach that subtly disturbs key parameters in SSMs to implant backdoors. Specifically, we introduce three imperceptible periodic patterns as triggers and embed them into benign samples to generate poisoned samples. Our experiments show that BadVim is effective and successfully evades existing defenses, with hidden attention matrices resembling those of clean samples. We hope that our findings will inform the development of more robust defense strategies for VSSMs.

% Moreover, our findings show that while the SSM mechanism enhances model capabilities, it also introduces significant risks to backdoor robustness. 

%% file: sec/X_suppl.tex
% \clearpage
% \setcounter{page}{1}
% \setcounter{section}{0}
% \renewcommand{\thesection}{\Alph{section}}
% \maketitlesupplementary

The content of Supplementary Material is organized as follows: (1) Section~\ref{abl:s4} reviews the State Space Model (SSM) paradigm to explain its application in VSSMs; (2) Section~\ref{abl:badvim} and \ref{abl:proof} describe the BadVim algorithms in detail and provide analytical evidence to elucidate their impact on VSSMs; (3) Section~\ref{abl:expt-set} presents the implementation and training details, including datasets, hyperparameters, and attack settings, to ensure the reproducibility of our work; (4) Section~\ref{abl:results} and \ref{abl:insight} provide additional experimental results and discuss the differences between Gated CNNs and VSSMs.

\section{Revisit the concept of S4}\label{abl:s4}
We first define the HiPPO matrix~\citep{gu2020hippo}, which specifies a class of matrices $\mathbf{A} \in \mathbb{R}^{\mathsf{N} \times \mathsf{N}}$ that, when incorporated into Equation~\ref{Eq: SSM-Hidden}, enables the state $x(t)$ to retain the history of the input $u(t)$. The HiPPO matrix is given by
\begin{equation}\label{eq:hippo}
    \mathbf{A}_{nk} = -\begin{cases}
        (2n+1)^{1/2} (2k+1)^{1/2}, & \text{if } n>k,  \\
        n+1, & \text{if } n=k,  \\
        0, & \text{if } n<k.
    \end{cases}
\end{equation}

A key concept in S4~\citep{gu2022efficiently} is the efficient computation of all views of SSMs (Section~\ref{pre:ssm}). Specifically, S4 establishes an equivalence relation on SSMs: $(\mathbf{A}, \mathbf{B}, \mathbf{C}) \sim (\mathbf{V}^{-1} \mathbf{A} \mathbf{V}, \mathbf{V}^{-1} \mathbf{B}, \mathbf{C} \mathbf{V})$. This transformation accelerates computation when $\mathbf{A}$ is diagonal. Ideally, $\mathbf{A}$ would be diagonalizable by a unitary matrix, as guaranteed by the Spectral Theorem for normal matrices. However, normality is a restrictive condition that the HiPPO matrix does not satisfy. S4 observes that the HiPPO matrix can be decomposed into normal and low-rank components, though this alone does not enable efficient computation. To address this, S4 introduces three techniques (\textit{i.e}, Cauchy Kernel, Woodbury Identity, and truncated generating function) to overcome this bottleneck and generalize the Normal Plus Low-Rank (NPLR) decomposition to arbitrary matrices.
\begin{theorem}
All HiPPO matrices from~\citep{gu2020hippo} admit an NPLR decomposition of the form
\begin{equation}
\mathbf{A} = \mathbf{V} \Lambda \mathbf{V}^* - \mathbf{P} \mathbf{Q}^{\intercal}
= \mathbf{V} (\Lambda - (\mathbf{V}^* \mathbf{P})(\mathbf{V}^* \mathbf{Q})^) \mathbf{V}^,
\end{equation}
where $\mathbf{V} \in \mathbb{C}^{\mathsf{N} \times \mathsf{N}}$ is unitary, $\Lambda$ is diagonal, and $\mathbf{P}, \mathbf{Q} \in \mathbb{R}^{\mathsf{N} \times r}$ form a low-rank factorization. The HiPPO matrices LegS, LegT, and LagT satisfy $r = 1$ or $r = 2$. In particular, Equation~\ref{eq:hippo} corresponds to an NPLR decomposition with $r = 1$.
\end{theorem}

We omit variant HiPPO matrices (\textit{i.e}, LegS, LegT, and LagT) in this paper; please refer to S4~\citep{gu2022efficiently} for further details. Here, we note that any VSSM based on low-rank matrix approximation~\citep{clarkson2017low} can reduce the time complexity of state updates.
% markovsky2012low

\section{Details of BadVim}\label{abl:badvim}
\subsection{Trigger pattern}\label{abl:pattern}
In the BadVim framework, we present three strategies for constructing trigger patterns using various frequency- or pixel-based techniques. Notably, embedding the trigger pattern in each patch increases its robustness to perturbations. A straightforward approach is to use a static trigger, such as a single pixel, placed repeatedly across the image~\citep{shejwalkar2023perils}. This ensures that the pattern remains spatially consistent and unaffected by geometric transformations. Although similar in form to prior work~\citep{shejwalkar2023perils}, our method differs in both its goal and application context. Despite the state-sequence updates and patch-wise processing in VSSMs, we find that these embedded triggers can still successfully compromise the model.

\subsection{Our Algorithms}\label{abl:algo}
Our attacks are described in Algorithm~\ref{algo: BadVim}. Specifically, BadVim enables an adversary to select one of the strategies and inject it into the training data with a predefined poisoning rate. This approach highlights that when applied to each patch, low-rank perturbations can influence VSSMs during state updates and are more effective than SOTA attacks.

\input{sec/our_algo}

\section{Proofs}\label{abl:proof}
% We first introduce the following two assumptions commonly observed in encoders trained using SSL.

% \begin{theorem}\normalfont \label{thm:impact}
%     Let $\mathbf{M}$ be a state-space model (SSM) with a low-rank state-space matrix $\mathbf{A}$, such that $\mathbf{A} = \mathbf{U} \Lambda \mathbf{V}^{T}$, where $\mathbf{A}$ has rank $r \ll d$. Let $\Delta$ be a low-rank perturbation with $\text{rank}(\Delta) = r' \ll d$. Then, there exists a sequence of hidden states $\mathbf{H}_{t}$, governed by the recurrence relation:
%     \begin{equation*}
%         \mathbf{H}_{t}=(\mathbf{A}+\alpha \Delta)^{t} \mathbf{H}_{0}+\sum_{k=1}^{t}(\mathbf{A}+\alpha \Delta)^{k-1}\mathbf{B}\mathbf{X}_{t-k}.
%     \end{equation*}
%     Given that both $\mathbf{A}$ and $\Delta$ are low-rank matrices, their composition preserves the low-rank structure, ensuring that the perturbation $\Delta$ persists through multiple state updates. Thus, the model $\mathbf{M}$ retains the effects of low-rank triggers significantly better than high-rank noise.
% \end{theorem}

\begin{proof}[Proof of Proposition 1]
We first define the state evolution over multiple time steps under the presence of the perturbation during training and inference.

\noindent \textbf{Training Formulation.} Consider the training state evolution equation without perturbation: \[h_t=\mathbf{\overline{A}} h_{t-1} + \mathbf{\overline{B}} x_t.\] Now, suppose that the training samples are corrupted with a low-rank perturbation $\delta$, such that the poisoned input is $\hat{x}_t = x_t + \delta$. The corresponding state evolution equation becomes:
\[h_t' = \mathbf{\overline{A}}' h_{t-1}' + \mathbf{\overline{B}}' (x_t + \delta),\] 
where $\mathbf{\overline{A}}'$ and $\mathbf{\overline{B}}'$ denote the model parameters learned under the influence of the perturbation. Since the training data is modified, the learned transition matrix deviates from its clean counterpart:
$\mathbf{\overline{A}}' = \mathbf{\overline{A}} + E, \mathbf{\overline{B}}' = \mathbf{\overline{B}} + F,$ where $E$ and $F$ capture the differences induced by training on perturbed data. Notably, if the perturbation $\delta$ is low-rank, then 
$E$ is also expected to be low-rank due to its dependence on $\delta$.

\noindent \textbf{Propagating the State Evolution.} Expanding the recurrence over $t$ steps, we obtain:
\[h_t' = \mathbf{\overline{A}}^t h_0 + \sum_{i=0}^{t-1} \mathbf{\overline{A}}^i \mathbf{\overline{B}} \hat{x}_{t-i} \]
Similarly, under backdoor training:
\[h_t' = \mathbf{\overline{A}}'^t h_0 + \sum_{i=0}^{t-1} \mathbf{\overline{A}}'^i \mathbf{\overline{B}}' \hat{x}_{t-i} + \sum_{i=0}^{t-1} \mathbf{\overline{A}}'^i \mathbf{\overline{B}}' \delta. \]
The term $ S_t = \sum_{i=0}^{t-1} \mathbf{\overline{A}}'^i \mathbf{\overline{B}}' \delta $ captures the cumulative effect of the perturbation across time.

% which persists over training iterations. Given that $\delta$ is low-rank, it will propagate through the system, and its impact on the state will accumulate with training.

\noindent \textbf{Spectral Radius Condition.} To understand the long-term behavior of $S_t$, we analyze the spectral radius of $\mathbf{\overline{A}}'$. Applying standard spectral perturbation bounds:
\[\rho(\mathbf{\overline{A}'}) \geq \rho(\mathbf{\overline{A}}) - \|E\|.\]

Since the perturbation matrix $E$ is low-rank with a small norm, its effect on the spectral radius is negligible unless $\mathbf{\overline{A}}$ is already near a critical threshold. If $\rho(\mathbf{\overline{A}'}) \geq 1$, then small perturbations $E$ are insufficient to ensure contraction, implying $\rho(\mathbf{\overline{A}'}) \geq 1$ in most cases.

% We examine the effect of the state transition matrix $\mathbf{\overline{A}'}$ during the training phase with poisoned data. Since the perturbation propagates through $\mathbf{\overline{A}'}$, its long-term behavior depends on the spectral radius $\rho(\mathbf{\overline{A}'})$:

We now analyze the two cases:
\begin{itemize}
    \item \textbf{Case 1:} $\rho(\mathbf{\overline{A}'}) = 1$
    In this case, $\mathbf{\overline{A}'}^i$ does not decay. 
    So, the accumulated perturbation $S_t$ remains significant and persists throughout the state trajectory.
    
    \item \textbf{Case 2}: $\rho(\mathbf{\overline{A}'}) > 1$ The term $\mathbf{\overline{A}'}^i$ grows, leading to an unbounded accumulation of the perturbation effect. As a result, the poisoned model's state diverges significantly from the clean model over time.
     % The perturbation will grow over time. Since $\mathbf{\overline{A}'}$ is not decaying, the influence of the perturbation will increase over time. 
\end{itemize}
Since in both cases, the perturbation $\delta$ does not decay, we conclude that it persists in the model's state evolution. Under the condition $\rho(\mathbf{\overline{A}'}) \geq 1$, the perturbation's influence remains non-negligible throughout training and inference.

% Since $\rho(\mathbf{\overline{A}'}) \geq 1$ in the system, the low-rank perturbation $\delta$ persists in the state-space model and does not decay or disappear during training. Thus, the perturbation will continue to influence the state evolution for all $t$, and the effect of this perturbation will not disappear in inference.
\end{proof}

\begin{figure*}[!ht]
  \centering
  \includegraphics[width=\textwidth]{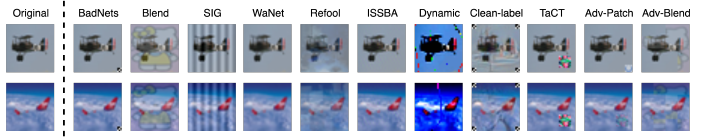}
  % \vspace{0.1mm}
  \caption{Example images of backdoored samples from CIFAR-10 dataset with 11 attacks.}
  \vspace{0.3cm}
  \label{fig:backdoor-example}
\end{figure*}

\section{Experimental Settings}\label{abl:expt-set}

\noindent \textbf{Datasets.}
We provide details of the datasets used in our experiments in Table~\ref{tab:details_dataset}, showing the number of classes, training samples, and test samples for each. These datasets~\citep{krizhevsky2009learning,stallkamp2011german,deng2009imagenet} are commonly used to evaluate backdoor attacks.

\begin{table}[!ht]
\begin{center}
\vspace{0.1mm}
\caption{Statistics of datasets used in our experiments.}
% \vspace{-0.15in}
\vspace{0.1cm}
\setlength{\tabcolsep}{2mm}
\resizebox{\columnwidth}{!}{%
\begin{tabular}{ccccc}
\toprule[0.68pt] \addlinespace[1.5pt]
Dataset & Input size & Classes & Training data & Test data \\  \hline \addlinespace[0.1em]
CIFAR10 & $3\times32\times32$ & 10 & 50,000 & 10,000 \\ 
% \hline \addlinespace[0.1em]
GTSRB & $3\times32\times32$ & 43 & 39,209 & 12,630 \\
% \hline \addlinespace[0.1em]
ImageNet & $3\times224\times224$ & 1000 & 1,281,167 & 100,000 \\
\addlinespace[-0.1em] \bottomrule[0.68pt]
\end{tabular}}\label{tab:details_dataset}
\end{center}
\end{table}

\noindent \textbf{Training Setting.}\label{abl:expt}
Following the training settings in Vim~\citep{zhu2024vision}, we adopt an AdamW optimizer with a momentum of $0.9$, a weight decay of $1 \times 10^{-8}$, a mixup rate of $0.8$, a warmup learning rate of $1 \times 10^{-5}$, and an initial learning rate of $5 \times 10^{-6}$ in our experiments. With a batch size of $128$, we fine-tune the pre-trained model on ImageNet for three epochs. Furthermore, we perform the experiments using a single NVIDIA V100 GPU with 32GB RAM, which requires approximately $16$ hours for the above experiments. Besides, for CIFAR10 and GTSRB, we set the patch size to $4$, the embedding dimension to $256$, and the model depth to $12$, in contrast to the settings of $16$, $192$, and $24$ used for ImageNet. For DeiT~\citep{touvron2021training} and GatedCNN~\citep{yu2025mambaout}, we follow the original setting in the official code to train models. Since the pre-trained models are only available for ImageNet, we train the models from scratch on CIFAR10 and GTSRB. 

\noindent \textbf{Attacks Setting.}
We consider several backdoor attacks and follow the default configuration in Backdoor-Toolbox~\citep{qi2023revisiting}. The attacks in our experiments have a poisoning rate of $0.3\%$. Furthermore, since Backdoor-ToolBox does not implement the clean-label attack (CLB) on GTSRB, we do not use this attack on GTSRB. Besides, our trigger pattern focuses on the large-scale dataset, aiming to enhance backdoor effectiveness. Consequently, we have omitted experiments on CIFAR10 and GTSRB. For clarity, we present visualization results of the backdoor attacks used in this paper, as shown in Figure~\ref{fig:backdoor-example}, using examples from CIFAR10. These include both visible and invisible types, as well as adaptive attacks~\citep{qi2023revisiting} such as Adv-Patch and Adv-Blend.

% \begin{table}
% \begin{subtable}{\columnwidth}\centering
% \resizebox{0.8\textwidth}{!}{
% \setlength{\tabcolsep}{1mm}
% {\begin{tabular}{ccccc}
%     \toprule
%     BadNets & Blend & SIG & Refool & TaCT\\
%     \midrule
%     1\% & 1\% & 2\% & 1\% & 2\%\\
%     \midrule
%     Dynamic & ISSBA & WaNet & Adv-Patch & Adv-Blend \\ 
%     \midrule
%     1\% & 2\% & 5\% & 1\% & 1\%  \\ 
%     \bottomrule
%     \end{tabular}}}
% \caption{CIFAR10}\label{tab:poison-rate-a}
% \end{subtable}
% \\[0.5em]
% \begin{subtable}{\columnwidth}\centering
% \resizebox{0.8\textwidth}{!}{
% \setlength{\tabcolsep}{1mm}
% {\begin{tabular}{ccccc}
%     \toprule
%     BadNets & Blend & SIG & Refool & TaCT \\
%     \midrule
%     1\% & 1\% & 2\% & 1\% & 0.5\% \\
%     \midrule
%     Dynamic & ISSBA & WaNet & Adv-Patch & Adv-Blend \\
%     \midrule
%     0.3\% & 2\% & 5\% & 0.5\% & 0.3\%\\ 
%     \bottomrule
%     \end{tabular}}}
% \caption{GTSRB}\label{tab:poison-rate-b}
% \end{subtable}
% \caption{Details of poisoning rate used in our paper.}\label{tab:poison-rate}
% \end{table}

\noindent \textbf{Defenses Setting.}\label{abl:defense}
Since existing defenses are primarily designed for CNN and ViT architectures, we adapt five model-agnostic backdoor defenses: two for patch processing and three for in-training defense.

\begin{figure}[!h]
    \centering
    \includegraphics[width=0.8\columnwidth]{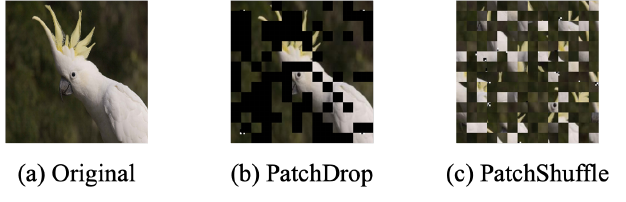}
    \vspace{0.1mm}
    \caption{Illustration of patch processing defense. We randomly drop patches or shuffle the patches within the image to disrupt trigger patterns, with a drop rate of $0.3$ and patch size of $16$ in this sample.}
    \vspace{0.1cm}
    \label{fig:patch_defense}
\end{figure}

\noindent \textbf{(1.) Patch processing defense:} Recall that patch processing techniques~\citep{doan2023defending} effectively disrupt trigger patterns to combat backdoor attacks on ViTs. Specifically, the first approach, PatchDrop~\citep{naseer2021intriguing}, randomly removes a certain number of patches from an image, leading to information loss. The second approach, PatchShuffle~\citep{kang2017patchshuffle}, randomly shuffles the patches within the image's spatial grid. Despite the fact that this method retains the image content, it significantly impacts the receptive fields of the models. To ensure effectiveness, we set the drop rate as $0.3$ in PatchDrop~\citep{naseer2021intriguing} and used the model's original patch size for PatchShuffle~\citep{kang2017patchshuffle}, as shown in Figure~\ref{fig:patch_defense}. A higher drop rate and smaller patch size result in more severe disruption of spatial information.

\noindent \textbf{(2.) In-training defense:} We fine-tune with auxiliary clean data (\textit{i.e}, $5\%$ of training data). FT~\citep{liu2018fine} and FT-SAM~\citep{zhu2023enhancing} are trained for $100$ epochs. FT-SAM uses a sharpness-aware optimizer to minimize the loss function, thereby enhancing model generalization by reducing the sharpness of the loss surface, whereas FT applies traditional fine-tuning without such enhancements. For FT, we perform fine-pruning by removing $20\%$ of the weights in the convolutional and linear layers before fine-tuning for $10$ epochs.

\section{More Experimental Results}\label{abl:results}

% \noindent \textbf{Results of Loss Changes.}
% In Fig. xxx, we present the loss changes of Vim across different triggers during fine-tuning. Generally, effective backdoors lead to faster convergence in the model, resulting in lower overall loss. As observed, the pre-trained model fine-tuned on clean data exhibits a relatively higher loss. In contrast, the models with backdoors show lower losses, with our backdoor achieving the lowest. This supports the effectiveness of our main paper as stated.
  
% \noindent \textbf{Results of Clean Accuracy.}
% As shown in Fig. xxx, fine-tuning the pre-trained model on clean data achieves higher accuracy compared to the models trained with the three types of poisoned data. Although the backdoored models do not match the accuracy of the clean model, the decreases are negligible (less than $1$\%). The clean accuracy of our proposed method is between BadNets and Blend.

% \noindent \textbf{Results of Attack Success Rate.}
% Fig. xxx shows the results of backdoor attacks. As for fine-tuning the pre-trained model on ImageNet, we find that BadNets achieves an ASR of over $99$\%, while Blend approaches $82$\%. This indicates that the VSSM remains susceptible to patch-wise triggers. However, as we mentioned, patch-wise triggers are ineffective after patch processing. Our trigger pattern not only overcomes this defense but also maintains effectiveness comparable to BadNets.

\noindent \textbf{Different strategies in BadVim.}\quad Table~\ref{tab:diff-stra} shows that all three BadVim strategies achieve high ASR on CIFAR10 and GTSRB. Notably, the ``\textit{LRP-Checkerboard}'' pattern attains $100$\% ASR on both datasets while maintaining strong ACC values of $85.58$\% and $93.65$\%, respectively. In comparison, the ``\textit{LRP-Sine}'' and ``\textit{LRP-Stripes}'' trigger patterns yield slightly lower ASR and ACC, with the ``\textit{LRP-Sine}'' pattern achieving $96.32$\% ASR on CIFAR10 and $86.48$\% on GTSRB. These results further corroborate our finding that the state transitions inherent to VSSMs make it prone to backdoor attacks. Therefore, we adopt the ``\textit{LRP-Checkerboard}'' pattern for the main results in this paper.

% In Table~\ref{tab:diff-stra}, we illustrate the backdoor robustness between two models by varying the pixel intensity from $8$ to $30$ under an extreme poisoning rate of $0.3$\%. As anticipated, increasing the trigger pixel intensity enhances the attack's effectiveness. Even at an intensity of $10$, our attack remains effective at implanting backdoors in Vim, whereas MambaOut shows a lower ASR (\textit{i.e.}, $32.35$\%). This further supports our findings on the impact of the SSM mechanism in the VSS model on backdoor robustness. Importantly, increasing pixel intensity in our trigger does not affect the ACC.

\input{tab/tab8-diff-strategies}
\vspace{-0.1in}

\noindent \textbf{Evaluation of Trigger Pattern Quality.}\quad Table~\ref{tab:abl-perturb} summarizes the image quality after applying different low-rank perturbation (LRP) trigger strategies. Among the three designs, the checkerboard pattern achieves the highest PSNR (\textit{i.e.}, $37.27$ on CIFAR10, $37.38$ on GTSRB) and SSIM scores (\textit{i.e.}, $0.9893$ and $0.9781$, respectively), indicating minimal visual distortion when embedding the trigger. In contrast, the sine and stripe patterns yield lower PSNR and SSIM values, reflecting more noticeable perturbations. These results suggest that the ``\textit{LRP-Checkerboard}'' trigger achieves a better trade-off between stealthiness and attack effectiveness across both datasets.

\input{tab/tab9-quality}

% \begin{figure*}[!ht]
%     \centering
%     \begin{minipage}{0.32\textwidth}
%         \centering
%         \includegraphics[width=0.8\columnwidth]{Figure/png/loss-epoch.png}
%         \caption{Loss changes on different backdoors during fine-tuning.}
%         \label{fig:loss-epoch}
%     \end{minipage}\hfill
%     \begin{minipage}{0.32\textwidth}
%         \centering
%         \includegraphics[width=0.8\columnwidth]{Figure/png/acc-epoch.png}
%         \caption{Accuracy changes on different backdoors during fine-tuning.}
%         \label{fig:acc-epoch}
%     \end{minipage}\hfill
%     \begin{minipage}{0.32\textwidth}
%         \centering
%         \includegraphics[width=0.8\columnwidth]{Figure/png/asr-epoch.png}
%         \caption{ASR changes on different backdoors during fine-tuning.}
%         \label{fig:asr-epoch}
%     \end{minipage}
%     % \caption{xxxxx}
%     \label{fig:analysis}
% \end{figure*}

% As shown in Fig.~\ref{fig:acc-epoch}, fine-tuning the pre-trained model on clean data achieves significantly higher accuracy compared to the models trained with the three types of poisoned data. Although the backdoored models do not match the accuracy of the clean model, the decreases are negligible (less than 1\%). The clean accuracy of our proposed is between BadNets and Blend.

\section{Distinction between Gated CNNs and VSSMs}\label{abl:insight}
\begin{figure}[!h]
  \centering
  \includegraphics[width=0.75\columnwidth]{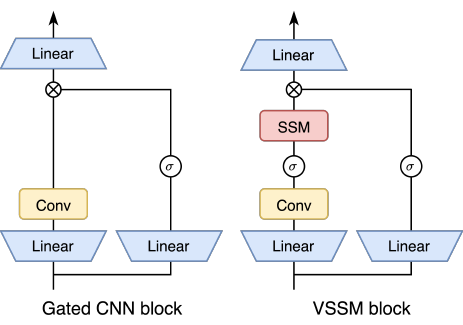}
  \vspace{0.1cm}
  \caption{Difference between Gated CNNs and VSSM block. For simplicity, we omit the normalization and the shortcut.}
  \vspace{0.2in}
  \label{fig:model-arch}
\end{figure}

Recent studies~\citep{yu2025mambaout} argue that visual models do not require a causal mode (\textit{i.e.}, the SSM mechanism) for effective task understanding, as most classification datasets (\textit{e.g.}, small-scale images) lack long-sequence dependencies. In response, they propose GatedCNN, a model composed of stacked Gated CNN blocks (see Figure~\ref{fig:model-arch}), which consistently outperforms VSSMs across various image sizes in classification tasks. Building on prior architectures such as ConvNeXt~\citep{liu2022convnet} and InceptionNeXt~\citep{yu2024inceptionnext}, GatedCNN achieves both simplicity and strong performance. The key distinction between VSSMs and GatedCNN lies in their design: while VSSMs depend on sequence memorization, stacked CNN blocks in GatedCNN offer greater versatility and efficiency in feature extraction.

Based on our preceding discussion, we summarize our observations regarding the necessity of SSM robustness against backdoor attacks as follows:

\begin{itemize}
    \item ACC robustness on benign data: VSSM $<$ Gated CNNs
    \item ASR sensitivity: VSSM $>>$ Gated CNNs
    \item Gap between ACC and ASR: VSSM $<$ Gated CNNs
\end{itemize}

%%%%%    others    %%%%%

% \begin{figure}[!t]
%   \centering
%   \centerline{\includegraphics[width=0.85\columnwidth]{Figure/pdf/Backdoor sample.pdf}}
%   \caption{Visualization of the clean and backdoor samples, with the trigger location marked with a red frame.}\label{fig:backdoor}
% \end{figure}

% 分析程度 (Which attack sensitive?) → Explanation (SSM角度)
% 說明SSM對於Patch-based較Blend-based(whole image)的ASR來的高許多，並解釋...

% 由前子章節得到的結論，並且根據MambaOut結構及文獻進行分析解釋
% 說明/解釋為什麼會具有Robusteness/穩健性
% providing distinct advantages in processing long sequences and compressing them into a constant memory size.
% significantly enhances robustness exhibits strong learning capability even when only a few poisoned samples are present in the training set, with a relatively small drop in clean accuracy.

% \begin{figure}[!t]
%   \centering
%   \centerline{\includegraphics[width=\columnwidth]{fig/pdf/Illustration.pdf}}
%   \caption{Illustration of patch processing and the VSS model.}
%   \Description{The left side demonstrates the PatchDrop (drop rate = 0.2) and PatchShuffle (size = 16) techniques used in our experiments. The right side depicts the VSS model component, which splits images into patches, projects them into patch tokens, and sends the sequence of tokens to the VSS encoder. In this paper, we use Vim for the VSS model.}\label{fig:illustation}
% \end{figure}

%% file: sec/our_algo.tex
\newcommand\mycommfont[1]{\footnotesize\ttfamily\textcolor{blue}{#1}}
\SetCommentSty{mycommfont}

\begin{algorithm}[!hbp]
    \DontPrintSemicolon
	\caption{Training of BadVim}
	\label{algo: BadVim}
	\textbf{Input}: Training data $D_{train} = \{(x_i, y_i)\}$, a loss function $\mathcal{L}_{vss}$, a mapping function $\mathcal{T}$, three strategies $\mathcal{S} = \{S_1, S_2, S_3\}$, where $S_1$ is ``Sinusoidal wave,'' $S_2$ is ``Checkerboard,'' and $S_3$ is ``Stripes.'' In addition, let $f_x/f_y$ denote the frequency intensity and $\alpha$ the blending rate.
 
    \textbf{Output}: Backdoored Model $\mathcal{M}_{bd}$.

    \tcc{Step 0: Pick a strategy}
    $S^\star \leftarrow \mathcal{S}$
    
    \tcc{$D_{train}$ is a training dataset}
    \For{$(x_i, y_i) \in D_{train}$}{

        \tcc{Step 1: Generate BadVim trigger based on the selected strategy}
        \If{$S^\star == S_1$}{
            \tcc{Generate sinusoidal wave trigger (LRP-Sine)}
            $I(x, y) = \sum_k [a_k \sin(2\pi \omega_k x + \phi_k) + b_k \sin(2\pi \omega_k y + \psi_k)]$
        }
        \ElseIf{$S^\star == S_2$}{
            \tcc{Generate checkerboard trigger (LRP-Checkerboard)}
            $I(x, y) = \operatorname{sign}(\sin(2\pi f_x x) \sin(2\pi f_y y))$
        }
        \ElseIf{$S^\star == S_3$}{
            \tcc{Generate stripe trigger (LRP-Stripes)}
            $I(x, y) = \operatorname{sign}(\sin(2\pi f_x x) + \sin(2\pi f_y y))$
        }
    
    \tcc{Step 2: Apply the trigger to the pixel domain and blend it with the clean image}
    $x_i' = \mathcal{T}(x_i, I(x, y), \alpha)$ 
    
    \tcc{Step 3: Update the training dataset with the poisoned images}
    $D_{train}^{bd} \leftarrow D_{train} \cup \{x_i'\}$
    }
    \tcc{Train a backdoored model $\mathcal{M}_{bd}$ on the poisoned data}
    Randomly initialize a model $\mathcal{M}$ \\
    $\mathcal{M}_{bd} = \mathcal{M}(D_{train}^{bd}, \mathcal{L}_{vss})$
    \Return Model $\mathcal{M}_{bd}$
\end{algorithm}

%% file: tab/tab8-diff-strategies.tex
\begin{table}[!ht]
\begin{center}
\vspace{0.1mm}
\caption{Backdoor attack performance across three strategies.}
% \vspace{-0.15in}
\vspace{0.1cm}
\setlength{\tabcolsep}{1mm}{
\resizebox{0.75\columnwidth}{!}{%
\begin{tabular}{c|cc|cc|cc}
\toprule
Strategy $\rightarrow$ &\multicolumn{2}{|c|}{LR-Sine} & \multicolumn{2}{|c}{LR-Check} & \multicolumn{2}{|c}{LR-Stripes} \\ \midrule
Dataset $\downarrow$ & ACC & ASR & ACC & ASR & ACC & ASR \\ \midrule
CIFAR10 & \underline{85.23} & 96.32 & \textbf{85.58} & \textbf{100} & 84.88 & \underline{97.56} \\ 
GTSRB & 92.28 & 86.48 & 93.65 & \textbf{100} & \underline{93.42} & \underline{98.89} \\ \bottomrule
\end{tabular}}}\label{tab:diff-stra}
\end{center}
\end{table}

%% file: tab/tab9-quality.tex
\setlength{\tabcolsep}{2pt}
\begin{table}[!ht]
\begin{center}
\vspace{0.1mm}
\caption{Comparison of image quality with our three strategies on CIFAR10 and GTSRB.}
\vspace{0.1cm}
\scalebox{0.95}{
\begin{tabular}{lcccc}
\toprule
\multirow{2}{*}{Strategy} & \multicolumn{2}{c}{CIFAR10} & \multicolumn{2}{c}{GTSRB}\\
\cmidrule{2-5}
& PSNR & SSIM & PSNR & SSIM \\
\midrule
LRP-Sine & 34.23 & $0.9652{\pm.026}$ & 34.26 & $0.9310{\pm.058}$\\
LRP-Checkerboard & 37.27 & $0.9893{\pm.009}$ & 37.38 & $0.9781{\pm.020}$\\
LRP-Stripes & 32.60 & $0.9831{\pm.015}$ & 32.71 & $0.9645{\pm.033}$\\
\bottomrule
\end{tabular}}\label{tab:abl-perturb}
\end{center}
\end{table}

%% file: reference.bib
@inproceedings{he2016deep,
  title={Deep residual learning for image recognition},
  author={He, Kaiming and Zhang, Xiangyu and Ren, Shaoqing and Sun, Jian},
  booktitle={CVPR},
  pages={770--778},
  year={2016},
}

@inproceedings{papernot2016limitations,
  title={The limitations of deep learning in adversarial settings},
  author={Papernot, Nicolas and McDaniel, Patrick and Jha, Somesh and Fredrikson, Matt and Celik, Z Berkay and Swami, Ananthram},
  booktitle={IEEE EuroS\&P},
  pages={372--387},
  volume={},
  number={},
  year={2016},
  organization={IEEE}
}

@inproceedings{
dosovitskiy2021an,
title={An Image is Worth 16x16 Words: Transformers for Image Recognition at Scale},
author={Alexey Dosovitskiy and Lucas Beyer and Alexander Kolesnikov and Dirk Weissenborn and Xiaohua Zhai and Thomas Unterthiner and Mostafa Dehghani and Matthias Minderer and Georg Heigold and Sylvain Gelly and Jakob Uszkoreit and Neil Houlsby},
booktitle={ICLR},
year={2021},
}

@inproceedings{gu2023mamba,
  title={Mamba: Linear-time sequence modeling with selective state spaces},
  author={Gu, Albert and Dao, Tri},
  booktitle={COLM},
  year={2024}
}

@inproceedings{zhu2024vision,
  title={Vision Mamba: Efficient Visual Representation Learning with Bidirectional State Space Model},
  author={Lianghui Zhu and Bencheng Liao and Qian Zhang and Xinlong Wang and Wenyu Liu and Xinggang Wang},
  booktitle={ICML},
  year={2024},
}

@inproceedings{
dao2024transformers,
title={Transformers are {SSM}s: Generalized Models and Efficient Algorithms Through Structured State Space Duality},
author={Tri Dao and Albert Gu},
booktitle={ICML},
year={2024},
}

@inproceedings{
liu2024vmamba,
title={{VM}amba: Visual State Space Model},
author={Yue Liu and Yunjie Tian and Yuzhong Zhao and Hongtian Yu and Lingxi Xie and Yaowei Wang and Qixiang Ye and Jianbin Jiao and Yunfan Liu},
booktitle={NIPS},
year={2024},
}

@inproceedings{li2024mamba,
  title={Mamba-nd: Selective state space modeling for multi-dimensional data},
  author={Li, Shufan and Singh, Harkanwar and Grover, Aditya},
  booktitle={ECCV},
  pages={75--92},
  year={2024},
  organization={Springer}
}

@inproceedings{wu2024rainmamba,
title={RainMamba: Enhanced Locality Learning with State Space Models for Video Deraining},
author={Hongtao Wu and Yijun Yang and Huihui Xu and Weiming Wang and Jinni Zhou and Lei Zhu},
booktitle={ACM MM},
year={2024},
}

@inproceedings{
zhang2024voxel,
title={Voxel Mamba: Group-Free State Space Models for Point Cloud based 3D Object Detection},
author={Guowen Zhang and Lue Fan and Chenhang He and Zhen Lei and Zhaoxiang Zhang and Lei Zhang},
booktitle={NIPS},
year={2024},
}

@inproceedings{xing2024segmamba,
title = {SegMamba: Long-range Sequential Modeling Mamba For 3D Medical Image Segmentation},
author = {Xing, Zhaohu and Ye, Tian and Yang, Yijun and Liu, Guang and Zhu, Lei},
booktitle = {MICCAI},
year = {2024},
page = {578 -- 588}
}

@inproceedings{
wang2024the,
title={The Mamba in the Llama: Distilling and Accelerating Hybrid Models},
author={Junxiong Wang and Daniele Paliotta and Avner May and Alexander M Rush and Tri Dao},
booktitle={NIPS},
year={2024},
}

@inproceedings{hatamizadeh2024mambavision,
  title={Mambavision: A hybrid mamba-transformer vision backbone},
  author={Hatamizadeh, Ali and Kautz, Jan},
  booktitle={CVPR},
  year={2025}
}

@article{du2024understanding,
  title={Understanding robustness of visual state space models for image classification},
  author={Du, Chengbin and Li, Yanxi and Xu, Chang},
  journal={arXiv preprint arXiv:2403.10935},
  year={2024}
}

@inproceedings{yu2025mambaout,
  title={MambaOut: Do We Really Need Mamba for Vision?},
  author={Yu, Weihao and Wang, Xinchao},
  booktitle={CVPR},
  year={2025}
}

@inproceedings{gu2022efficiently,
title={Efficiently Modeling Long Sequences with Structured State Spaces},
author={Albert Gu and Karan Goel and Christopher Re},
booktitle={ICLR},
year={2022},
}

@inproceedings{smithsimplified,
  title={Simplified State Space Layers for Sequence Modeling},
  author={Smith, Jimmy TH and Warrington, Andrew and Linderman, Scott},
  booktitle={ICLR},
  year={2023},
}

@inproceedings{fuhungry,
  title={Hungry Hungry Hippos: Towards Language Modeling with State Space Models},
  author={Fu, Daniel Y and Dao, Tri and Saab, Khaled Kamal and Thomas, Armin W and Rudra, Atri and Re, Christopher},
  booktitle={ICLR},
  year={2023},
}

@inproceedings{mehta2023long,
title={Long Range Language Modeling via Gated State Spaces},
author={Harsh Mehta and Ankit Gupta and Ashok Cutkosky and Behnam Neyshabur},
booktitle={ICLR},
year={2023},
}

@article{weng2020trade,
  title={On the trade-off between adversarial and backdoor robustness},
  author={Weng, Cheng-Hsin and Lee, Yan-Ting and Wu, Shan-Hung Brandon},
  journal={NIPS},
  volume={33},
  pages={11973--11983},
  year={2020}
}

@article{qiu2023towards,
  title={Towards a critical evaluation of robustness for deep learning backdoor countermeasures},
  author={Qiu, Huming and Ma, Hua and Zhang, Zhi and Abuadbba, Alsharif and Kang, Wei and Fu, Anmin and Gao, Yansong},
  journal={IEEE TIFS}, 
  year={2023},
  volume={19},
  number={},
  pages={455-468},
  publisher={IEEE}
}

@inproceedings{yuan2023you,
  title={You Are Catching My Attention: Are Vision Transformers Bad Learners under Backdoor Attacks?},
  author={Yuan, Zenghui and Zhou, Pan and Zou, Kai and Cheng, Yu},
  booktitle={CVPR},
  pages={24605--24615},
  year={2023}
}

@inproceedings{doan2023defending,
  title={Defending backdoor attacks on vision transformer via patch processing},
  author={Doan, Khoa D and Lao, Yingjie and Yang, Peng and Li, Ping},
  booktitle={AAAI},
  year={2023}
}

@article{touvron2022resmlp,
  title={Resmlp: Feedforward networks for image classification with data-efficient training},
  author={Touvron, Hugo and Bojanowski, Piotr and Caron, Mathilde and Cord, Matthieu and El-Nouby, Alaaeldin and Grave, Edouard and Izacard, Gautier and Joulin, Armand and Synnaeve, Gabriel and Verbeek, Jakob and others},
  journal={IEEE TPAMI}, 
  volume={45},
  number={4},
  pages={5314--5321},
  year={2022},
  publisher={IEEE}
}

@inproceedings{subramanya2024closer,
  title={A Closer Look at Robustness of Vision Transformers to Backdoor Attacks},
  author={Subramanya, Akshayvarun and Koohpayegani, Soroush Abbasi and Saha, Aniruddha and Tejankar, Ajinkya and Pirsiavash, Hamed},
  booktitle={WACV},
  pages={3874--3883},
  year={2024}
}

@article{gu2019badnets,
  title={Badnets: Evaluating backdooring attacks on deep neural networks},
  author={Gu, Tianyu and Liu, Kang and Dolan-Gavitt, Brendan and Garg, Siddharth},
  journal={IEEE Access},
  volume={7},
  pages={47230--47244},
  year={2019},
  publisher={IEEE}
}

@article{li2022backdoor,
  title={Backdoor learning: A survey},
  author={Li, Yiming and Jiang, Yong and Li, Zhifeng and Xia, Shu-Tao},
  journal={IEEE TNNLS}, 
  year={2022},  
  volume={35},
  number={1},
  pages={5-22},
  publisher={IEEE}
}

@article{nguyen2020input,
  title={Input-aware dynamic backdoor attack},
  author={Nguyen, Tuan Anh and Tran, Anh},
  journal={NIPS},
  volume={33},
  pages={3454--3464},
  year={2020}
}

@inproceedings{li2021invisible,
  title={Invisible backdoor attack with sample-specific triggers},
  author={Li, Yuezun and Li, Yiming and Wu, Baoyuan and Li, Longkang and He, Ran and Lyu, Siwei},
  booktitle={ICCV},
  pages={16463--16472},
  year={2021}
}

@inproceedings{tang2021demon,
  title={Demon in the variant: Statistical analysis of $\{$DNNs$\}$ for robust backdoor contamination detection},
  author={Tang, Di and Wang, XiaoFeng and Tang, Haixu and Zhang, Kehuan},
  booktitle={USENIX},
  pages={1541--1558},
  year={2021}
}

@article{turner2019label,
  title={Label-consistent backdoor attacks},
  author={Turner, Alexander and Tsipras, Dimitris and Madry, Aleksander},
  journal={arXiv preprint arXiv:1912.02771},
  year={2019}
}

@inproceedings{liu2020reflection,
  title={Reflection backdoor: A natural backdoor attack on deep neural networks},
  author={Liu, Yunfei and Ma, Xingjun and Bailey, James and Lu, Feng},
  booktitle={ECCV},
  pages={182--199},
  year={2020},
  organization={Springer}
}

@article{li2020invisible,
  title={Invisible backdoor attacks on deep neural networks via steganography and regularization},
  author={Li, Shaofeng and Xue, Minhui and Zhao, Benjamin Zi Hao and Zhu, Haojin and Zhang, Xinpeng},
  journal={IEEE TDSC},
  volume={18},
  number={5},
  pages={2088--2105},
  year={2020},
  publisher={IEEE}
}

@inproceedings{zhao2022defeat,
  title={Defeat: Deep hidden feature backdoor attacks by imperceptible perturbation and latent representation constraints},
  author={Zhao, Zhendong and Chen, Xiaojun and Xuan, Yuexin and Dong, Ye and Wang, Dakui and Liang, Kaitai},
  booktitle={CVPR},
  pages={15213--15222},
  year={2022}
}

@inproceedings{barni2019new,
  title={A new backdoor attack in cnns by training set corruption without label poisoning},
  author={Barni, Mauro and Kallas, Kassem and Tondi, Benedetta},
  booktitle={IEEE ICIP},
  pages={101--105},
  year={2019},
  organization={IEEE}
}

@inproceedings{huang2022backdoor,
title={Backdoor Defense via Decoupling the Training Process},
author={Kunzhe Huang and Yiming Li and Baoyuan Wu and Zhan Qin and Kui Ren},
booktitle={ICLR},
year={2022},
}

@inproceedings{zhu2023enhancing,
  title={Enhancing fine-tuning based backdoor defense with sharpness-aware minimization},
  author={Zhu, Mingli and Wei, Shaokui and Shen, Li and Fan, Yanbo and Wu, Baoyuan},
  booktitle={ICCV},
  pages={4466--4477},
  year={2023}
}

@article{zhu2024neural,
  title={Neural polarizer: A lightweight and effective backdoor defense via purifying poisoned features},
  author={Zhu, Mingli and Wei, Shaokui and Zha, Hongyuan and Wu, Baoyuan},
  journal={NIPS},
  volume={36},
  year={2024}
}

@inproceedings{liu2018fine,
  title={Fine-pruning: Defending against backdooring attacks on deep neural networks},
  author={Liu, Kang and Dolan-Gavitt, Brendan and Garg, Siddharth},
  booktitle={RAID},
  pages={273--294},
  year={2018},
  organization={Springer}
}

@inproceedings{liu2022convnet,
  title={A convnet for the 2020s},
  author={Liu, Zhuang and Mao, Hanzi and Wu, Chao-Yuan and Feichtenhofer, Christoph and Darrell, Trevor and Xie, Saining},
  booktitle={CVPR},
  pages={11976--11986},
  year={2022}
}

@inproceedings{yu2024inceptionnext,
  title={Inceptionnext: When inception meets convnext},
  author={Yu, Weihao and Zhou, Pan and Yan, Shuicheng and Wang, Xinchao},
  booktitle={CVPR},
  pages={5672--5683},
  year={2024}
}

@inproceedings{qi2023revisiting,
title={Revisiting the Assumption of Latent Separability for Backdoor Defenses},
author={Xiangyu Qi and Tinghao Xie and Yiming Li and Saeed Mahloujifar and Prateek Mittal},
booktitle={ICLR},
year={2023},
}

@article{clarkson2017low,
  title={Low-rank approximation and regression in input sparsity time},
  author={Clarkson, Kenneth L and Woodruff, David P},
  journal={Journal of ACM},
  volume={63},
  number={6},
  pages={1--45},
  year={2017},
  publisher={ACM New York, NY, USA}
}

@article{krizhevsky2009learning,
  title={Learning multiple layers of features from tiny images},
  author={Krizhevsky, Alex and Hinton, Geoffrey and others},
  journal={Technical report, Univ. of Toronto},
  year={2009},
  publisher={Toronto, ON, Canada}
}

@inproceedings{stallkamp2011german,
  title={The German traffic sign recognition benchmark: a multi-class classification competition},
  author={Stallkamp, Johannes and Schlipsing, Marc and Salmen, Jan and Igel, Christian},
  booktitle={IJCNN},
  pages={1453--1460},
  year={2011}
}

@inproceedings{deng2009imagenet,
  title={Imagenet: A large-scale hierarchical image database},
  author={Deng, Jia and Dong, Wei and Socher, Richard and Li, Li-Jia and Li, Kai and Fei-Fei, Li},
  booktitle={CVPR},
  pages={248--255},
  year={2009}
}

@inproceedings{touvron2021training,
  title={Training data-efficient image transformers \& distillation through attention},
  author={Touvron, Hugo and Cord, Matthieu and Douze, Matthijs and Massa, Francisco and Sablayrolles, Alexandre and J{\'e}gou, Herv{\'e}},
  booktitle={ICML},
  pages={10347--10357},
  year={2021}
}

@article{chen2017targeted,
  title={Targeted backdoor attacks on deep learning systems using data poisoning},
  author={Chen, Xinyun and Liu, Chang and Li, Bo and Lu, Kimberly and Song, Dawn},
  journal={arXiv preprint arXiv:1712.05526},
  year={2017}
}

@inproceedings{nguyen2021wanet,
title={WaNet - Imperceptible Warping-based Backdoor Attack},
author={Tuan Anh Nguyen and Anh Tuan Tran},
booktitle={ICLR},
year={2021}
}

@inproceedings{shejwalkar2023perils,
  title={The perils of learning from unlabeled data: Backdoor attacks on semi-supervised learning},
  author={Shejwalkar, Virat and Lyu, Lingjuan and Houmansadr, Amir},
  booktitle={ICCV},
  pages={4730--4740},
  year={2023}
}

@article{naseer2021intriguing,
  title={Intriguing properties of vision transformers},
  author={Naseer, Muhammad Muzammal and Ranasinghe, Kanchana and Khan, Salman H and Hayat, Munawar and Shahbaz Khan, Fahad and Yang, Ming-Hsuan},
  journal={NIPS},
  volume={34},
  pages={23296--23308},
  year={2021}
}

@article{kang2017patchshuffle,
  title={Patchshuffle regularization},
  author={Kang, Guoliang and Dong, Xuanyi and Zheng, Liang and Yang, Yi},
  journal={arXiv preprint arXiv:1707.07103},
  year={2017}
}

@inproceedings{gao2019strip,
  title={Strip: A defence against trojan attacks on deep neural networks},
  author={Gao, Yansong and Xu, Change and Wang, Derui and Chen, Shiping and Ranasinghe, Damith C and Nepal, Surya},
  booktitle={ACSAC},
  pages={113--125},
  year={2019}
}

@inproceedings{huang2023distilling,
title={Distilling Cognitive Backdoor Patterns within an Image},
author={Hanxun Huang and Xingjun Ma and Sarah Monazam Erfani and James Bailey},
booktitle={ICLR},
year={2023}
}

@article{ali2024hidden,
  title={The hidden attention of mamba models},
  author={Ali, Ameen and Zimerman, Itamar and Wolf, Lior},
  journal={arXiv preprint arXiv:2403.01590},
  number={},
  volume={},
  year={2024}
}

@inproceedings{han2024demystify,
  title={Demystify Mamba in Vision: A Linear Attention Perspective},
  author={Han, Dongchen and Wang, Ziyi and Xia, Zhuofan and Han, Yizeng and Pu, Yifan and Ge, Chunjiang and Song, Jun and Song, Shiji and Zheng, Bo and Huang, Gao},
  booktitle={NIPS},
  year={2024},
}

@article{gu2020hippo,
  title={Hippo: Recurrent memory with optimal polynomial projections},
  author={Gu, Albert and Dao, Tri and Ermon, Stefano and Rudra, Atri and R{\'e}, Christopher},
  journal={NIPS},
  volume={33},
  pages={1474--1487},
  year={2020}
}

@inproceedings{ravanelli2018speaker,
  title={Speaker recognition from raw waveform with sincnet},
  author={Ravanelli, Mirco and Bengio, Yoshua},
  booktitle={IEEE spoken language technology workshop (SLT)},
  pages={1021--1028},
  year={2018},
  organization={IEEE}
}

@inproceedings{
cirone2024theoretical,
title={Theoretical Foundations of Deep Selective State-Space Models},
author={Nicola Muca Cirone and Antonio Orvieto and Benjamin Walker and Cristopher Salvi and Terry Lyons},
booktitle={NIPS},
year={2024},
}

@inproceedings{
qi2024exploring,
title={Exploring Adversarial Robustness of Deep State Space Models},
author={Biqing Qi and Yiang Luo and Junqi Gao and Pengfei Li and Kai Tian and Zhiyuan Ma and Bowen Zhou},
booktitle={NIPS},
year={2024},
}
